# New Mathematical and Algorithmic Schemes for Pattern Classification with Application to the Identification of Writers of Important Ancient Documents

D. Arabadjis[1], F. Giannopoulos[1], C. Papaodysseus[1], S. Zannos[1], P. Rousopoulos[2], M. Panagopoulos[3] and C. Blackwell[4]

1. National Technical University of Athens, Department of Electrical and Computer Engineering, 9 Heroon Polytechneiou, GR-15773, Athens, GREECE.
2. Technical Institute of Chalkida, Department of Automatic Control, Chalkida, Euboea, GREECE.
3. Ionian University, Department of Audio & Visual Arts, Corfu, GREECE.
4. Furman University, Department of Classics, Greenville, South Carolina, U.S.A.

Corresponding author: Constantin Papaodysseus[1], cpapaod@cs.ntua.gr, tel. (+30) 210 772 1476, (+30) 210 361 7438

## Abstract

In this paper, a novel approach is introduced for classifying curves into proper families, according to their similarity. First, a mathematical quantity we call plane curvature is introduced and a number of propositions are stated and proved. Proper similarity measures of two curves are introduced and a subsequent statistical analysis is applied. First, the efficiency of the curve fitting process has been tested on 2 shapes datasets of reference. Next, the methodology has been applied to the very important problem of classifying 23 Byzantine codices and 46 Ancient inscriptions to their writers, thus achieving correct dating of their content. The inscriptions have been attributed to ten individual hands and the Byzantine codices to four writers.

# 1. Introduction

## 1.1 The importance of identifying the writer of ancient inscriptions and Byzantine codices

The set of surviving handwritten documents is one of the main sources for the science of History. More specifically, carved in stone, ancient inscriptions are the most important means of studying Antiquity. Similarly, manuscripts written on both papyri and parchments contributed to the transmission of the ancient world's literature through the Middle Ages, finally leading to the Renaissance and the Enlightenment. For example, the Homeric Iliad survives mainly through a handful of large manuscript volumes, all produced in Constantinople during the 10th or 11th century and currently scattered in different libraries throughout Europe: Venice, El Escorial in Spain, London, Geneva, Florence and Rome. These volumes contain the Homer's poem itself, as well as a number of different commentary texts and short notes in the margins of the manuscript and between lines. As one can easily assume from the above, dating of the content of these inscriptions and manuscripts is of particularly high significance for both disciplines of History and Archaeology. "Proper historical use of inscriptions can only be made if they can be dated", as stated by one of the most influential historians, Prof. Christian Habicht. However, the writers of ancient inscriptions and manuscripts rarely signed or dated their documents, making the process of dating them really difficult; this fact often causes disputes and disagreements among scientists. One major goal of the present paper is to quantitatively analyze the content of a given set of ancient inscriptions and Byzantine codices, so as to determine their relative dates of production, as well as the relationship among them. In fact, carving inscriptions was a profession in Antiquity. The working careers of most ancient writers covered about 20 to 25 years, while very few worked for 40 years, at most. If one achieves in attributing the ancient inscriptions to their writer, then evidently, one has also successfully dated their content. Similar arguments hold for Byzantine codices, which as a rule, were reproduced by monks of the era.

## 1.2 The goal of the present work

The present paper tries to tackle three problems: 1) To develop and present a new method for optimally matching two curves, one of which may be subject to two independent scaling transformations (along either x or y-axis), rotation and parallel translation. 2) To introduce proper, novel statistical criteria, in order to classify a given ensemble of such curves into proper families/clusters, according to their similarity, based on introduced, new similarity measures. 3) To classify a set of important ancient inscriptions and byzantine codices to the proper writer, so that these documents can be unambiguously dated.

There is an underlying fundamental assumption in the proposed solution of the aforementioned problems, which we will elucidate in the case of writer identification: We assume that when a specific writer generates a realization of an alphabet symbol on a document, then the writer may alter the orientation, position and size of the produced letter, rather arbitrarily; however, still, there is a kernel in the generated realization, which remains invariant under the aforementioned transformations and, in addition, is peculiar to the writer himself. Evidently, this, also, may hold true in relation to many other procedures/human activities, as for example in pottery, in painting and arts in general, in contour distortion by noise, etc.

## 1.3     A brief state of the art in matching and grouping planar shapes

Shapes comparison in [1] is treated as a features matching over orientation and geometric characteristics evaluated by differential properties of the implicit shape's function. Then, grouping of shapes is performed in a decision trees – based hierarchical clustering context. A features vector – based shape categorization approach is formulated in [2], in a statistical manner. The features of shapes that form a group are mapped on the linear base of their Gaussian graded – mean covariance matrix, which is selected to be the shapes group representative. Registration of planar shapes under affine transformations is studied in [3] using signed Euclidean distance implicit representations of planar shapes. The matching problem is then formulated via the maximization of the mutual information metric in the space of the parameters of affine transformation; maximization is achieved via a steepest descent method. Orientation invariant comparison of planar shapes is treated in [4] in terms of a "proper" comparison between the sequences of their curvature values. Propriety of the comparison is determined as the re-parameterization of the curves that mostly benefits curvatures correspondence, optimized over all possible differential point-to-point correspondences, using Dijkstra's algorithm. The approach introduced in [5] concerns N-dimensional curve registration and comparison by evaluating the alignment of the tangential directions of a curve to another fixed curve's tangents in the conjugate gradient context. In [6] comparison of (closed) planar shapes is based on prototyping shapes deviations as if they were caused by a Newtonian vector flow acting in the interior of each shape. Then, one-to-one shape correspondence is determined as geodesic paths minimizing deformation strains. In [7] and [8], deformation vector flow is modeled as an infinite dimensional (Hilbert) manifold over 1D functions descriptive of closed curves modulo scaling and modulo rotation (in [7]). Then, the curves fitting error is selected so as to retain the inner product defined on the manifold of the integral constraints; this allows for projection and translation of the functional representation of the shapes along paths that minimize the chosen error. A similar formulation that overcomes the infinite expansion of the manifold's directions is given in [9], where the 2D orthogonal expansion of the shapes' functional representation is guaranteed by embedding directional vectors in the complex plain. Then, a mapping is constructed, so as to convert the selected metric space into a Euclidean one and thus to "linearize" the problem of geodesics. Authors of [10] extend the representation of a shape in the form of a linear combination of 1D functions to the 2D case, using DC functions (i.e. the difference of 2 convex functions). This is achieved by solving a corresponding $L^1$ optimization problem, modulo uniform scaling and Euclidean transformations.

### 1.4 A brief state of the art in automatic writer identification

Recently, there has been a considerable interest in research on the topics of automated writer identification and verification, mainly concerning hand written text. Thus, various approaches have been developed, like the [11, 12, 13, 14, 15], based on feature extraction, while Hidden Markov Models are applied to [16, 17, 18]. A lot of research is also being undertaken in association with morphological approaches [19] or texture identification [20]. In [21] a Fourier

Transform approach of the pen-point movement barycenter's velocity is proposed and in [22] a dichotomy transformation is performed. [23] measures the individuality of handwritten characters through a number of identification and verification models, whereas connected-component contours and allograph prototype methods are described in [24, 25]. More recently, researchers in [26, 27] use a continuous character prototype distribution approach with fuzzy c-means algorithm in order to estimate the probability that a character has been generated by a prototype. Other scholars [28, 29] tackle the same problem by using a combination of local descriptors and learning techniques or by using directional morphological features [30]. Furthermore, others associate biometric and personal features with the handwriting style [31, 32, 33], while, in [34] certain characteristics of graphological type, such as skew, slant, pressure, thinning area, etc. are employed in order to classify calligraphic handwritten scripts according to their writer. Most recently, methods of automatic writer identification have been applied to text that consists of non – Latin symbols [35, 36], as well as to documents of historical importance [37, 38, 39].

## 2.     A number of fundamental definitions

A first major goal of the present work is to achieve optimal fit for an ensemble of curves. Towards this direction, we will first define a new quantity we call "plane curvature". In practice, all considered curves will be digital and will lie in a digital image; nevertheless, the strict approach will refer to continuous curves and, next, the algorithmic will consider finite affine and spatial steps. As it will become evident from Sect. 5.1 and the explicit form of the algorithm presented in Sect. 5, optimality of the continuous approach will ensure natural similarity of the involved digitized shapes.

**DEFINITION 1:** Consider a continuous, smooth Jordan curve $\Gamma_1$ embedded in a sub-domain $I$ of $R^2$, where $I$ plays the role of the image frame to-be-digitalized. Let $(x, y)$ be the coordinate system of the sub-domain $I$. Then, we define an implicit representation of $\Gamma_1$ as a zero isocontour of a twice differentiable function $F(x, y) : I \to R$, i.e. $F(x, y) = 0$ on $\Gamma_1$.

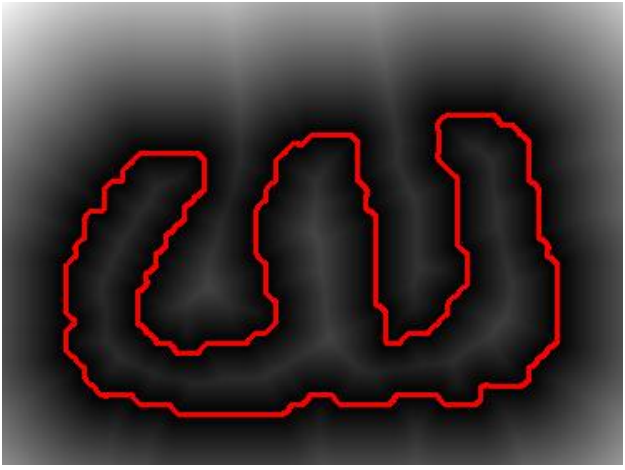

**Fig. 1** Depiction of the isocontours of the Euclidean distance from the contour of an alphabet symbol "omega". This letter contour, equivalently the zero-level isocontour, is shown in red.

**DEFINITION 2:** Let $E$ be an arbitrary equilevel curve (or isocontour) of $F(x, y) = 0$, i.e. the locus of points $(x, y) \in I$ such that $F(x, y) = c$, $c$ an arbitrary constant (see figure 1) and let $M$ be an arbitrary point of $E$.

Moreover, let $\vec{n}$ be the unit vector normal to $E$ at $M$, such that $\vec{n} = \dfrac{\nabla F}{|\nabla F|}$. Then, we define the function

$$C(x, y) \equiv \nabla \cdot \vec{n} \equiv \operatorname{div}(\vec{n}) \qquad (2.1)$$

where, $\nabla = \vec{i} \dfrac{\partial}{\partial x} + \vec{j} \dfrac{\partial}{\partial y}$ is the gradient operator. We will use the term "plane curvature of $F$" for function $C(x, y)$.

Since, curve $\Gamma_1$ is an isocontour of $F(x, y)$, with $c = 0$, $C(x, y)$ on $\Gamma_1$ is the actual standard curvature of $\Gamma_1$ at each point of it.

Now, suppose that a second smooth Jordan curve $\Gamma_2$ is also embedded in $I$ and let $\Omega$ be the sub-domain of $I$ included between $\Gamma_1$ and $\Gamma_2$ (see figure 2a). Then, we may express the similarity of curves $\Gamma_1$ and $\Gamma_2$ by means of the "plane curvature error function"

$$\varepsilon^C = \int_{\Omega} |\nabla \cdot \vec{n}| \, d\Omega \qquad (2.2)$$

namely, the double integral of the norm of $\operatorname{div}(\vec{n})$ on $\Omega$ (see figure 2b). Clearly, if curves $\Gamma_1$ and $\Gamma_2$ are identical and optimally fit, then $\varepsilon^C = 0$; if $\Gamma_2$ manifests deviations from $\Gamma_1$ and, still, the two curves are optimally fit, then $\varepsilon^C$ grows according to the degree of the deviations, as we will show in Sect. 5.1. In other words, the error function defined in (2.2) is strongly connected with a natural interpretation of the variation and similarity between $\Gamma_1$ and $\Gamma_2$. In the following, we will prove that if $\Gamma_2$ is subject to a proper set of transformations, i.e. suitable scaling, rotation and parallel displacement, then curves $\Gamma_1$ and $\Gamma_2$ can be optimally matched by minimization of $\varepsilon^C$.

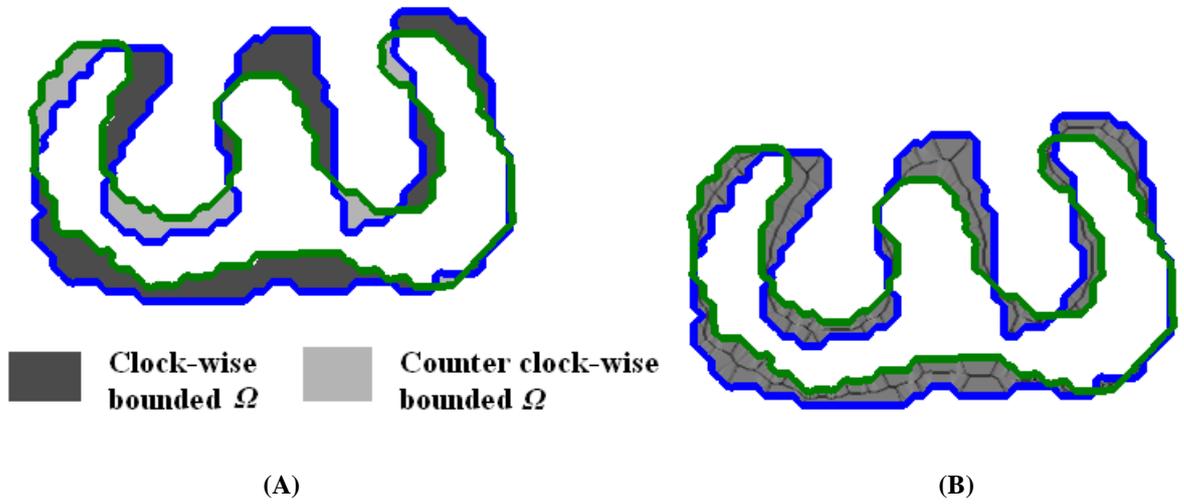

**(A)**

**(B)**

**Fig. 2** The sub-domain $\Omega$ of discrepancy of the two curves of interest and the plane curvature error function.

**A**) The sub-domain $\Omega$ as formed by the discrepancy of the contours of two "omega" symbol realizations. The contour of reference $\Gamma_1$ is depicted in blue, whereas the transformed one $\Gamma_2$ is shown in green. Path $\Gamma_1 \rightarrow \Gamma_2$ defines the orientation of the boundary of $\Omega$ as depicted in the figure.

**B**) Plane curvature error function values in the sub-domain $\Omega$ shown in gray-scale; the lower the error value, the darker is the shade.

## 3.    A number of propositions concerning plane curvature

We may consider $\varepsilon^C$ as a functional, in the sense that each curve $\Gamma_2$ embedded in $I$ generates a different value of $\varepsilon^C$. Therefore, this functional may become minimum for a certain position of curve $\Gamma_2$, with respect to the fixed curve $\Gamma_1$; for this position the value of $\varepsilon^C$ will be stationery and, hence, $\delta(\varepsilon^C) = 0$, namely, the variation of $\varepsilon^C$ will be zero.

**PROPOSITION 1:** Error functional $\varepsilon^C$ is minimum if the following error functional

$$\zeta^C = \int_\Omega (\nabla \cdot \vec{n})^2 \, d\Omega \qquad (3.1)$$

is, also, minimum in the same position of curve $\Gamma_2$. Equivalently, $\delta(\varepsilon^C) = 0 \Leftrightarrow \delta(\zeta^C) = 0$.

Hence, the following holds

**PROPOSITION 2:** Minimization of error $\varepsilon^C$ is equivalently expressed by means of minimization of the following error function:

$$\eta^C = 2\oint_{\Gamma_2} \nabla \cdot \vec{n} \, dl \qquad (3.2)$$

*PROOF*

We will demonstrate the proposition by minimizing the functional $\zeta^C$. In fact, by applying Stokes Theorem, we obtain:

$$\int_\Omega (\nabla \cdot \vec{n})^2 \, d\Omega = \int_\Omega (\nabla \cdot \vec{n})(\nabla \cdot \vec{n}) d\Omega = \oint_{\partial\Omega} (\nabla \cdot \vec{n}) \, \vec{n} \cdot \hat{\kappa} \, dr - \int_\Omega \vec{n} \cdot (\nabla(\nabla \cdot \vec{n})) d\Omega \qquad (3.3)$$

where, $\partial\Omega$ is the boundary of the domain $\Omega$ and $\hat{\kappa}$ its unit normal vector.

But, $\vec{n} \cdot (\nabla(\nabla \cdot \vec{n}))$ is the directional derivative $\dfrac{\partial(\nabla \cdot \vec{n})}{\partial \vec{n}}$ along the unit vector $\vec{n}$. In addition, if $\vec{l}$ is the unit vector tangent to $\Gamma_2$ at an arbitrary point of it, $dl$ the elementary arclength of $\Gamma_2$ and $dn$ the elementary length normal to $\vec{l}$ towards $\Gamma_1$, then $d\Omega = -dn \, dl$; the reason for the minus sign is that we want curve $\Gamma_2$ to "move" towards $\Gamma_1$, opposite to the unit vector normal to $\partial\Omega$ of Stokes theorem.

Then, $\displaystyle\int_\Omega \vec{n} \cdot (\nabla(\nabla \cdot \vec{n})) d\Omega = \int_\Omega \dfrac{\partial(\nabla \cdot \vec{n})}{\partial \vec{n}} d\Omega = -\int_\Omega \dfrac{\partial(\nabla \cdot \vec{n})}{\partial \vec{n}} dn \, dl = -\oint_{\partial\Omega} \nabla \cdot \vec{n} \, dl$

Therefore, substituting in (3.3) we obtain:

$$\zeta^C = 2\oint_{\partial\Omega} \nabla \cdot \vec{n} \, dl \Leftrightarrow \zeta^C = 2\left( \oint_{\Gamma_2} \nabla \cdot \vec{n} \, dl - \oint_{\Gamma_1} \nabla \cdot \vec{n} \, dl \right) \qquad (3.4)$$

Functional $\zeta^C$ remains always greater than or equal to zero, hence, it has a minimum for a certain position of $\Gamma_2$ at which it holds $\delta(\zeta^C) = 0$. But, since curve $\Gamma_1$ is kept fixed, it follows that $\delta(\oint_{\Gamma_1} \nabla \cdot \vec{n}\, dl) = 0$. Hence, $\delta(\zeta^C) = 0 \Leftrightarrow \delta(\oint_{\Gamma_2} \nabla \cdot \vec{n}\, dl) = 0$

Taking into consideration proposition 1, too, it follows that $\delta(\varepsilon^C) = 0 \Leftrightarrow \delta(\oint_{\Gamma_2} \nabla \cdot \vec{n}\, dl) = 0$

*Q.E.D.*

**PROPOSITION 3:** Let $(x, y)$ be an arbitrary point of $I$, $E$ be the isocontour of $F(x, y)$ passing from it and $\vec{n} = n_x \vec{i} + n_y \vec{j}$ be the unit normal vector defined in Section 2.

Passing in matrix notation, let $\mathbf{\nabla} = \begin{bmatrix} \dfrac{\partial}{\partial x} \\ \dfrac{\partial}{\partial y} \end{bmatrix}$ be the gradient matrix, $\mathbf{H} = \begin{bmatrix} \dfrac{\partial}{\partial x} \\ \dfrac{\partial}{\partial y} \end{bmatrix} \begin{bmatrix} \dfrac{\partial}{\partial x} & \dfrac{\partial}{\partial y} \end{bmatrix} =$

$= \begin{bmatrix} \dfrac{\partial^2}{\partial x^2} & \dfrac{\partial^2}{\partial y \partial x} \\ \dfrac{\partial^2}{\partial x \partial y} & \dfrac{\partial^2}{\partial y^2} \end{bmatrix}$ be the matrix of the Hessian operator.

Then $\mu^2 = \begin{bmatrix} \dfrac{\partial F}{\partial x} & \dfrac{\partial F}{\partial y} \end{bmatrix} \begin{bmatrix} \dfrac{\partial F}{\partial x} \\ \dfrac{\partial F}{\partial y} \end{bmatrix} = \left(\dfrac{\partial F}{\partial x}\right)^2 + \left(\dfrac{\partial F}{\partial y}\right)^2$ (3.5)

is the norm of $\nabla F$; moreover, let $\mathbf{J} = \begin{bmatrix} 0 & 1 \\ -1 & 0 \end{bmatrix}$ be an auxiliary matrix. Then, the plane curvature at point $(x, y)$ is also expressed via the relation

$$C(x, y) = \dfrac{1}{\mu^3} \begin{bmatrix} \dfrac{\partial F}{\partial x} & \dfrac{\partial F}{\partial y} \end{bmatrix} \mathbf{J}^T \mathbf{H} \mathbf{J} \begin{bmatrix} \dfrac{\partial F}{\partial x} \\ \dfrac{\partial F}{\partial y} \end{bmatrix}$$ (3.6)

The proof of Proposition 3 is given in Appendix A

# 4. Plane curvature variation under affine transformations

To achieve optimal matching between curves $\Gamma_2$ and $\Gamma_1$, we apply the following original approach: We impose scaling, rotation and parallel displacement to the entire domain I and, in particular, to both curves $\Gamma_2$ and $\Gamma_1$. In this way, after a set of such transformations, we obtain the transformed curves $\widetilde{\Gamma}_2$ and $\widetilde{\Gamma}_1$. As a result, the isocontours of $\widetilde{\Gamma}_1$ also change, thus changing the value and direction of unit vector $\vec{n}$. Hence, plane curvature $C(x, y)$ also changes value and the purpose of the present Section is to calculate the precise form of this change. We would like to emphasize that although the entire domain I is subjected to the above affine transformations, nevertheless, the fitting errors $\zeta^C$, $\varepsilon^C$, $\eta^C$ are computed in the domain $\Omega$ and each contour $\partial\Omega$ defined by the non-transformed curve $\Gamma_1$ and the transformed one $\widetilde{\Gamma}_2$. Equivalently, we transform the entire domain $I$, in order to achieve proper affine transformations of $\Gamma_2$, but the resulting, under these transformations, curve $\widetilde{\Gamma}_2$ is always compared with the intact curve $\Gamma_1$ which is supposed to remain fixed over the transformed domain $I$.

The implementation of the approach requires stating and demonstrating the following set of definitions and propositions, associated with the change of $I$ under these transformations.

**DEFINITION 3:** We express the elementary scaling, rotation and parallel displacement imposed to every point of $I$, by means of the elementary matrix:

$$\mathbf{A} = \begin{bmatrix} 1+da & 0 \\ 0 & 1+db \end{bmatrix}\begin{bmatrix} 1 & -dT \\ dT & 1 \end{bmatrix} = \begin{bmatrix} 1 & 0 \\ 0 & 1 \end{bmatrix} + \begin{bmatrix} da & -dT \\ dT & db \end{bmatrix} \equiv I_2 + d\mathbf{A} \qquad (4.1)$$

where, we consider only first-order calculus and a mono-parametric group of transformations, $I_2$ is the 2x2 unit matrix, $d\mathbf{A}$ is the elementary part of $\mathbf{A}$, $(1+da)$ is the scaling factor along $x$-axis, $(1+db)$ is the scaling factor along $y$-axis and $dT$ is the elementary angle of rotation around the understood $z$-axis. If we also consider the elementary displacement $\begin{bmatrix} d\gamma_x \\ d\gamma_y \end{bmatrix}$, which

we assume it is the same for all points of domain $I$, then the arbitrary point $(x, y) \in I$, under

the set of these elementary transformations originates to a point $(\tilde{x}, \tilde{y})$ by the relation:

$$\begin{bmatrix} x \\ y \end{bmatrix} = \mathbf{A} \begin{bmatrix} \tilde{x} \\ \tilde{y} \end{bmatrix} + \begin{bmatrix} d\gamma_x \\ d\gamma_y \end{bmatrix} \qquad (4.2)$$

We should note that $(x, y)$ and $(\tilde{x}, \tilde{y})$ are in an one-to-one correspondence, since $\mathbf{A}$ is locally invertible with $\mathbf{A}^{-1} = I_2 - d\mathbf{A}$ in a first order approximation. Thus $(\tilde{x}, \tilde{y})$ results from the

infinitesimal transformation of $(x, y)$ by:

$$\begin{bmatrix} \tilde{x} \\ \tilde{y} \end{bmatrix} = (I_2 - d\mathbf{A}) \left( \begin{bmatrix} x \\ y \end{bmatrix} - \begin{bmatrix} d\gamma_x \\ d\gamma_y \end{bmatrix} \right)$$

**LEMMA 1:** The gradient operator $\mathbf{\nabla}$, under transformation (4.2), changes according to

$$\begin{bmatrix} \dfrac{\partial}{\partial \tilde{x}} \\ \dfrac{\partial}{\partial \tilde{y}} \end{bmatrix} = \mathbf{A}^T \begin{bmatrix} \dfrac{\partial}{\partial x} \\ \dfrac{\partial}{\partial y} \end{bmatrix} \Leftrightarrow \tilde{\mathbf{\nabla}} = \mathbf{A}^T \mathbf{\nabla} \qquad (4.3)$$

In addition, the norm of $\nabla F$ changes according to

$$\tilde{\mu} = \sqrt{\nabla^T F \, \mathbf{A} \mathbf{A}^T \nabla F} \qquad (4.4)$$

while, the Hessian operator is transformed according to

$$\tilde{\mathbf{H}} = \mathbf{A}^T \mathbf{H} \mathbf{A} \qquad (4.5)$$

**PROPOSITION 4:** Under transformations (4.2), the plane curvature changes by $dC$ given by

$$dC = C \, \frac{g_a da + g_b db}{\mu^2} \qquad (4.11)$$

where $g_a = 3 \left( \dfrac{\partial F}{\partial x} \right)^2 - 2\mu^2$ \qquad (4.11a)

and $g_b = 3 \left( \dfrac{\partial F}{\partial y} \right)^2 - 2\mu^2$ \qquad (4.11b)

*PROOF*

According to proposition 3, the plane curvature $\widetilde{C}(\widetilde{x}, \widetilde{y})$ is written in the $\widetilde{x}, \widetilde{y}$ coordinates as

$$\widetilde{C}(\widetilde{x}, \widetilde{y}) = \frac{\widetilde{\nabla}^T \widetilde{F} \, \mathbf{J}^T \widetilde{\mathbf{H}} \mathbf{J} \, \widetilde{\nabla} \widetilde{F}}{\widetilde{\mu}^3} \tag{4.12}$$

Using the results of Lemma 1, (4.12) is written:

$$\widetilde{C}(\widetilde{x}, \widetilde{y}) = \frac{\nabla^T F \, \mathbf{A} \mathbf{J}^T \mathbf{A}^T \mathbf{H} \, \mathbf{A} \mathbf{J} \mathbf{A}^T \, \nabla F}{(\nabla^T F \, \mathbf{A} \mathbf{A}^T \, \nabla F)^{3/2}} \tag{4.13}$$

But, $\mathbf{A} \mathbf{J}^T \mathbf{A}^T = \begin{bmatrix} 0 & -(1+da)(1+db) \\ (1+da)(1+db) & 0 \end{bmatrix} = (1+da)(1+db)\mathbf{J}^T \tag{4.14}$

Therefore, (4.13) now becomes:

$$\widetilde{C}(\widetilde{x}, \widetilde{y}) = (1+da)^2(1+db)^2 \frac{\nabla^T F \, \mathbf{J}^T \mathbf{H} \mathbf{J} \, \nabla F}{(\nabla^T F \, \mathbf{A} \mathbf{A}^T \, \nabla F)^{3/2}}$$

$$\Leftrightarrow \widetilde{C}(\widetilde{x}, \widetilde{y}) = (1+da)^2(1+db)^2 \frac{\nabla^T F \, \mathbf{J}^T \mathbf{H} \mathbf{J} \, \nabla F}{\mu^3} \frac{\mu^3}{(\nabla^T F \, \mathbf{A} \mathbf{A}^T \, \nabla F)^{3/2}}$$

$$\Leftrightarrow \widetilde{C}(\widetilde{x}, \widetilde{y}) = (1+da)^2(1+db)^2 \frac{\mu^3}{(\nabla^T F \, \mathbf{A} \mathbf{A}^T \, \nabla F)^{3/2}} C(x, y) \tag{4.15}$$

Letting $\widetilde{C} = C + dC$ and using (3.4), we obtain:

$$dC = C \left( \frac{(1+da)^2(1+db)^2 \, |\nabla F|^3 - (\nabla^T F \, \mathbf{A} \mathbf{A}^T \, \nabla F)^{3/2}}{(\nabla^T F \, \mathbf{A} \mathbf{A}^T \, \nabla F)^{3/2}} \right) \tag{4.16}$$

Now, $(\nabla^T F \, \mathbf{A} \mathbf{A}^T \, \nabla F)^{3/2} = \left( \left( \frac{\partial F}{\partial \widetilde{x}} \right)^2 + \left( \frac{\partial F}{\partial \widetilde{y}} \right)^2 + 2da \left( \frac{\partial F}{\partial \widetilde{x}} \right)^2 + 2db \left( \frac{\partial F}{\partial \widetilde{y}} \right)^2 \right)^{3/2}$

and after using Taylor expansion and keeping first order terms only, we obtain:

$$(\nabla^T F \, \mathbf{A} \mathbf{A}^T \, \nabla F)^{3/2} = \frac{da \left( 3 \left( \frac{\partial F}{\partial x} \right)^2 - 2\mu^2 \right) + db \left( 3 \left( \frac{\partial F}{\partial y} \right)^2 - 2\mu^2 \right)}{\mu^2} \tag{4.17}$$

Evidently, using the definition of $g_a = 3\left(\dfrac{\partial F}{\partial x}\right)^2 - 2\mu^2$ and $g_b = 3\left(\dfrac{\partial F}{\partial y}\right)^2 - 2\mu^2$ as in (4.11),

the desired result follows immediately. *Q.E.D.*

**COROLLARY 1:** Suppose that we, sequentially, apply a set of transformations of type (4.2),

thus obtaining a finite composite transformation, which maps the initial curve $\Gamma_2$ to another curve $\widetilde{\Gamma}_2$. Let $(x, y)$ be an arbitrary point of $\Gamma_2$, which, under this finite composite transformation, moves to point $(\widetilde{x}, \widetilde{y}) \in \widetilde{\Gamma}_2$. It is understood that during this transformation, the pair of scaling factors $(a, b)$ change along a curve $\varDelta$ in their coordinate system, starting at identity point $(1,1)$. Then, the plane curvature $\widetilde{C}(\widetilde{x}, \widetilde{y})$ is related to $C(x, y)$ via formula

$$\frac{\widetilde{C}(\widetilde{x}, \widetilde{y})}{C(x, y)} = e^{\int_{\varDelta} \frac{g_a da + g_b db}{\mu^2}} \tag{4.18}$$

where the line integral in the exponent is computed upon path $\varDelta$.

In order to use solution (4.18) to parameterize the affine transformation of the $xy$ - plane by means of a desired curvature deformation, we need to evaluate the sensitivity of the line integral of (4.18) to the variations of path $\Delta$. In other words, we should evaluate the $2^{\text{nd}}$ order variation of $\ln \widetilde{C}(\widetilde{x}, \widetilde{y})$ under the infinitesimal transformation (4.1). This evaluation is performed in

Appendix D and the obtained results are summarized in the following proposition

**PROPOSITION 5**: Under infinitesimal transformation (4.2), the second order variation of the plane curvature function logarithm $\ln C(x, y)$ is independent of the absolute size of the path differential $(da, db)$ and depends only on the ratio $\gamma = \dfrac{da}{db}$. Namely, for $\lambda = ab$ and

$\sigma = \dfrac{1}{2}(a^2 - b^2)$ it holds that

$$\frac{\partial}{\partial \lambda} d \ln C \bigg|_{\substack{a=1 \\ b=1}} = 0, \qquad d^2 \ln C = d\sigma^2 \frac{\partial^2}{\partial \sigma^2} \ln C \bigg|_{\substack{a=1 \\ b=1}} \sim \mathbf{n}^T \begin{bmatrix} 0 & 1 \\ 1 & 0 \end{bmatrix} \mathbf{n} \left( \gamma - \frac{1}{\gamma} \right) \tag{4.19}$$

Hence, according to the aforementioned Proposition 5, the variation of the tangent of the integration path $\Delta$ at the origin depends only on the tangent's direction and not on its measure. Therefore, the determination of the optimal transformation path $\Delta$ that minimizes the plane curvature discrepancy between two curves, depends only on the direction of the tangent vector at the transformation's identity point (starting point of $\Delta$) and, consequently, it is parameterization invariant. So, this optimization can be performed independently of the $(da, db)$ - discretization size, making the whole $\Delta$ - interpolation procedure uniformly stable and robust.

The previous analysis indicates that plane curvature variation, under transformations (4.1), (4.2), depends only on two independent variables and in particular, the scaling factors $(a, b)$. However, the optimum placement of two curves $\Gamma_2$ and $\Gamma_1$ depends on three additional parameters corresponding to the rotation of the curve $\Gamma_2$ around the *z*-axis, as well as on the parallel displacement of $\Gamma_2$ along *x,y*-axes. Since, we have previously established that optimal scaling can be achieved independently of rotation and displacement, by means of the introduced plane curvature approach, then it is logical to expect that there are infinitely many positions of curve $\Gamma_2$, for which the plane curvature error $\zeta^C$ is minimum; these positions, obviously, correspond to the infinitely many rotations and displacements that may be applied to $\Gamma_2$. Nevertheless, we may adopt the quite standard criterion of actually optimal fit of $\widetilde{\Gamma}_2$ and $\Gamma_1$ inside domain *I*, which demands the integral of Euclidean distances of all points of $\widetilde{\Gamma}_2$ from $\Gamma_1$ to be minimum; then, we may employ this criterion together with the previously obtained results, in order to achieve best fit of $\widetilde{\Gamma}_2$ and $\Gamma_1$ inside domain *I*, when all transformations are taken into consideration. This will be the subject of the next Section.

# 5.    A new method and the corresponding algorithm for fitting two curves $\Gamma_1$ and $\Gamma_2$

In this Section, we will employ the previously obtained results and we will define a new similarity measure, in order to present the steps of a novel algorithm, which performs best fitting of any two Jordan curves $\Gamma_1$ and $\Gamma_2$ inside domain $I$. This novel approach may be embedded in any function minimization algorithm, such as the Nelder-Mead. The introduced curve fitting method consists of two stages: I) In the first stage, the optimal resize factors $a$ and $b$ are obtained, for which the overall plane curvature difference of $\Gamma_1$ and the transformed $\Gamma_2$ is minimum. Evaluating both these resize factors $a$ and $b$ independently is legitimate, since we have already proved in previous Sections that this procedure is independent of rotation and parallel translation. We should note here that determination of the optimal scaling factors $a$, $b$ through simple normalization of the curves length or of the area they enclose, etc., is legitimate / correct only in the case where the normalized curves coincide. As it will be shown in Sect. 5.1 by minimizing the curvature measure we obtain optimal scaling for an arbitrary pair of curves, independently of Euclidean transformations. II) In the second stage, we first define a proper similarity measure of $\Gamma_1$ and the optimally resized $\Gamma_2$ to incorporate rotation and parallel translation. Then, the method may employ any function minimization algorithm, in order to determine the optimal relative match of $\Gamma_1$ and the transformed $\Gamma_2$. These stages are described below.

## 5.1 Connection of the selected curvature measure with natural measures of shapes similarity

The implicit curvature measure introduced in Definition 2, Section 2, is strongly connected with the natural interpretation of the possible variations of a shape, i.e. the variations caused by vector field actions respecting a metric. Namely, considering the geodesics with tangents along vector field $\vec{n} \cdot \nabla$, we can evaluate the deviations from an isocontour $\Gamma_1$ of $F(x, y)$, inside the family of curves that $F(x, y)$ induces, by the geodesic length $\delta(\vec{p}_1, \vec{p}) = \int\limits_{0}^{F(\vec{p})} \dfrac{|dF|}{\|\nabla F\|}$, between a point $\vec{p}_1 \in \Gamma_1$ and an arbitrary point $\vec{p}$ of the $xy$ – plane. Then, letting $\vec{p}$ belong to a planar

curve $\Gamma_2$, we can evaluate the deviation of $\Gamma_2$ from $\Gamma_1$ by the Lagrangian integral

$D(\Gamma_1, \Gamma_2) = \int_{\Gamma_1} \delta(\vec{p}_1(s), \vec{p}) ds$, where $s \in [0, \tau]$ is the curve length parameterization of $\Gamma_1$.

Then, any transformation of the $xy$ – plane with infinitesimal length $dt$ moves $(x, y)$ to

$(x', y') = (x, y) + dt \, \vec{v}$, where $\vec{v}$ is an arbitrary unit vector indicating the direction of the

transformation of $(x, y)$. The effect of this transformation on $D(\Gamma_1, \Gamma_2)$ is given by the action of the corresponding vector field $\chi = \vec{v} \cdot \nabla$ on the integrands of $D(\Gamma_1, \Gamma_2)$. Namely,

$$\chi[D(\Gamma_1, \Gamma_2)] = \int_{\Gamma_1} \chi[\delta(\vec{p}_1(s), \vec{p})] ds + \delta(\vec{p}_1(s), \vec{p}) \, d\chi[s(\vec{p}_1)]$$

Evaluating separately the action of $\chi$ on $\delta$ and $ds$, we have

$$\chi[\delta(\vec{p}_1(s), \vec{p})] = (\vec{v} \cdot \vec{n})(\vec{p}) - (\vec{v} \cdot \vec{n})(\vec{p}_1), \qquad d(\chi[s]) = d(\vec{v} \cdot \vec{l})$$

where $\vec{l}$ is the unit vector tangent to $\Gamma_1$. Thus, the infinitesimal deviation of $D(\Gamma_1, \Gamma_2)$ reads

$$\chi[D(\Gamma_1, \Gamma_2)] = \left[ \delta(\vec{p}_1(s), \vec{p}) \, \vec{v}(\vec{p}) \cdot \vec{l}(\vec{p}_1) \right]_{s=0}^{s=\tau} + \int_{\Gamma_1} \left[ \vec{v} \cdot \vec{n} \right]_{\vec{p}_1(s)}^{\vec{p}} ds - \left[ d\delta \, \vec{v} \right]_{\vec{p}_1(s)}^{\vec{p}} \cdot \vec{l}$$

But, while moving on $\Gamma_1$ we always have $\delta(\vec{p}_1, \vec{p}_1) = 0$, implying that $d\delta(\vec{p}_1, \vec{p}_1) = 0$. In addition, moving on paths along $\vec{n}$ that connect points $\vec{p}_1$ and $\vec{p}$, $\vec{p}_1$ remains unchanged implying that $ds = 0$.

Thus, $\left[ \delta(\vec{p}_1, \vec{p}) \, \vec{v}(\vec{p}) \cdot \vec{l}(\vec{p}_1) \right]_{s=0}^{s=\tau} = \int_{\Gamma_1 \to \Gamma_2} \left[ d\delta \, \vec{v} \right]_{\vec{p}_1(0)}^{\vec{p}} \cdot \vec{l} + \int_{\Gamma_2 \to \Gamma_1} \left[ d\delta \, \vec{v} \right]_{\vec{p}_1(\tau)}^{\vec{p}} \cdot \vec{l}$ .

So, considering domain $\Omega$ defined by the paths connecting points $\vec{p}_1$ and $\vec{p}$ for all $\vec{p}_1(s) \in \Gamma_1$, $s \in [0, \tau]$, we can evaluate the 1st order variation of $D(\Gamma_1, \Gamma_2)$ using Stokes' theorem in the form

$$\chi[D(\Gamma_1, \Gamma_2)] = -\int_{\partial\Omega} \vec{v} \cdot \hat{\kappa} \, d\partial\Omega = -\int_{\Omega} \nabla \cdot \vec{v} \, d\Omega \qquad (5.1)$$

Where $\hat{\kappa}$ is the unit vector normal to $\partial\Omega$, with orientation that respects positive step $ds$ along $\Gamma_1$. Expressing differential displacement $dt\,\vec{v}$ of the Cartesian coordinates in the curvilinear frame $(\vec{n}, \vec{l})$, we obtain $dt\,\vec{v} = \vec{n}d\delta + \vec{l}\,ds$. Also, it holds that $\nabla \cdot (dt\,\vec{v}) = dt\nabla \cdot \vec{v} + \vec{v} \cdot \nabla dt$. Since $\vec{v}$ is unit vector and $dt$ differential length along $\vec{v}$, $\nabla dt = d\vec{v}$ and $\vec{v} \cdot d\vec{v} = 0$, thus implying that $\nabla \cdot (dt\,\vec{v}) = dt\nabla \cdot \vec{v}$. Evaluating $\nabla \cdot (d\delta\,\vec{n})$ and $\nabla \cdot (ds\,\vec{l})$ in the same way, we obtain $dt\nabla \cdot \vec{v} = d\delta\,\nabla \cdot \vec{n} + ds\,\nabla \cdot \vec{l} = -(ds - d\delta)\nabla \cdot \vec{n} \Rightarrow \nabla \cdot \vec{v} = -\vec{v} \cdot (\vec{l} - \vec{n})\nabla \cdot \vec{n}$.

Substituting this expression for $\nabla \cdot \vec{v}$ in (5.1), the 1st order variation of $D(\Gamma_1, \Gamma_2)$ has norm

$$\left\| \chi[D(\Gamma_1, \Gamma_2)] \right\| \leq \sqrt{2} \int_{\Omega} |\nabla \cdot \vec{n}| \, d\Omega \qquad (5.2)$$

which is the implicit curvature measure of (2.2). Therefore, the introduced implicit shapes similarity measure is the supremum of the Euclidean norm of the 1st order variation of the geodesic length between a pair of shapes, inside a given curves' family.

### 5.2 Optimal transforming $\Gamma_2$ in size, so as to fit $\Gamma_1$.

Matching $\Gamma_2$ to $\Gamma_1$, as far as minimum plane curvature error is concerned, comprises the following steps:

**Step 1**: We consider curve $\Gamma_2$ at its original position inside $I$ and for each point $(x, y) \in \Gamma_2$, we compute quantities $g_a, g_b, \mu$ via relations (4.11a), (4.11b), (3.5).

**Step 2**: We initially set $a=1$, $b=1$ so as to begin from the identity point of the transformation to be computed. Then:

a) We first compute $\varepsilon_1 = \oint_{\Gamma_1} \nabla \cdot \vec{n}\, dl$; this integral is numerically computed only once and on the nearest to $\Gamma_1$ isocontour, outwards to $\Gamma_2$, in the digital domain $I$.

b) At each point $(x, y)$ of $\Gamma_2$ we consider the isocontour of $F(x, y)$ passing from it, we compute $\nabla \cdot \vec{n}$ and we attribute this value to $C(x, y)$. This ensemble of values $C(x, y)$ play the role of the initial values of the plane curvature in curve $\Gamma_2$ and for this reason, we will, hereafter, employ for it the symbol $C_0^{\Gamma_2}(x, y)$. Evidently, subscript 0 indicates that this curvature is computed at position 0, namely the initial one, while superscript $\Gamma_2$ indicates that this curvature refers to $\Gamma_2$. We would like to point out that we actually obtain a set of values of $C_0^{\Gamma_2}(x, y)$, where each value is computed at the center of a corresponding pixel of the digitized version of $\Gamma_2$. In addition, we emphasize that $\nabla \cdot \vec{n}$ for $\Gamma_1$ is computed on the nearest isocontour outwards to it; we do so, in order to avoid discontinuities in the value of $\nabla \cdot \vec{n}$, on curve $\Gamma_1$, due to the fact that unit vector $\vec{n}$ changes orientation as we cross the corresponding curve.

**Step 3**: In order to decrease error function $|\zeta^C|$ under the infinitesimal deformation (4.11), we determine the locally optimal finite steps $\delta a, \delta b$ (discretizing d$a$,d$b$), in the frame of a function minimization algorithm, for example the Nelder-Mead. Next, the corresponding change of plane curvature $C_0^{\Gamma_2}(x, y)$ for all previously defined points in Step 2, is computed via (4.11) to give

$$dC(x, y) = C_0^{\Gamma_2}(x, y)\frac{g_a \delta a + g_b \delta b}{\mu^2}.$$ Then, the new plane curvature $C_1^{\Gamma_2}(x_1, y_1)$, at point $(x_1, y_1)$ obtained from $(x, y)$ via resize by factors $(1 + \delta a, 1 + \delta b)$, is

$C_1^{\Gamma_2}(x_1, y_1) = C_0^{\Gamma_2}(x, y) + dC(x, y)$. Performing this update for all appoints of $\Gamma_2$, we re-estimate (re-draw) it applying to all $(x, y) \in \Gamma_2$ the following transformation, which is actually

(4.1) and (4.2) properly adjusted: $\begin{bmatrix} x_1 \\ y_1 \end{bmatrix} = \left( \begin{bmatrix} 1 & 0 \\ 0 & 1 \end{bmatrix} + \begin{bmatrix} \delta a & 0 \\ 0 & \delta b \end{bmatrix} \right) \begin{bmatrix} x \\ y \end{bmatrix}$. We emphasize that in

(4.1) and (4.2), we have set the rotation and parallel displacement parameters equal to zero, i.e. $dT = d\gamma_x = d\gamma_y = 0$, since, as we have proved in the previous Sections, the change of plane curvature is independent of rotation and parallel translation. The ensemble of points $(x_1, y_1)$ forms the transformed version of digital curve $\Gamma_2$, say $\tilde{\Gamma}_2(x_1, y_1)$.

**Step 4**: The error functional $|\zeta^C|$ is computed for the new version $\tilde{\Gamma}_2(x_1, y_1)$ by means of relation (3.4). In fact, we digitally compute $\varepsilon_2 = \oint_{\Gamma_2} \nabla \cdot \vec{n} \, dl$ by numerically integrating $C_1^{\Gamma_2}(x_1, y_1)$ on $\tilde{\Gamma}_2$. If $|\zeta^C|$ is smaller than a properly selected threshold, usually associated with the function minimization algorithm, then the procedure stops. Otherwise, it proceeds to the next Step 5, which is actually Step 1 with different starting values.

**Step 5**: We compute $g_a, g_b, \mu$ via relations (4.11a), (4.11b), (3.5) on $\tilde{\Gamma}_2(x_1, y_1)$.

**Step 6**: We return to Step3 and we repeat the optimal transformation process up to Step 5 for $\tilde{\Gamma}_2(x_1, y_1)$, thus obtaining a new transformed version $\tilde{\Gamma}_2(x_2, y_2)$ of digital curve $\Gamma_2$ and so on, until the termination condition in Step 4 is met. Namely, until the error function $|\zeta^C|$ reaches a minimum.

At the end of this process, we obtain the best $\tilde{\Gamma}_2$ version of $\Gamma_2$, as far as difference of plane curvature of $\tilde{\Gamma}_2$ from $\Gamma_1$ is concerned ($|\zeta^C|$ is minimum). We should denote here that, in Sect. 4, by means of Proposition 5, it has been shown that the performance of this optimization procedure does not rely on the absolute size of the discretization steps $(\delta a, \delta b)$; this size only affects the density of the transformation's identity points' selection.

## 5.3 Incorporating rotation and parallel translation for optimally fitting $\Gamma_2$ to $\Gamma_1$.

From the moment that the best $\tilde{\Gamma}_2$ has been evaluated, one expects that there would be infinitely many different placements of $\tilde{\Gamma}_2$ relative to $\Gamma_1$, for which $|\zeta^C|$ is minimum, since, as we have already shown, minimization of plane curvature difference is independent of rotation and parallel translation. Therefore, the developed matching algorithm must account for these transformations, too. Thus, one may employ any typical error function and integrate it into a function minimization algorithm. For reasons that will be analytically explained in the following, we have chosen the subsequent, quite standard matching error:

Let $P$ be an arbitrary point of $\tilde{\Gamma}_2$ and let $d(P)$ be the Euclidean distance of $P$ from $\Gamma_1$. Then, the error of fitting $\tilde{\Gamma}_2$ to $\Gamma_1$ that has been employed is $\oint_{\tilde{\Gamma}_2} d(P)dl$, namely, the curvilinear integral of the distance of all points $P$ of $\tilde{\Gamma}_2$ from $\Gamma_1$.

At each step of the employed function minimization algorithm, the new position of $\tilde{\Gamma}_2$ is re-estimated by means of (4.1) and (4.2), after setting d$a$=d$b$=0, since the minimization of the plane curvature difference has already been achieved via the algorithm of Section 5.2. We note that rotation is applied only to $\tilde{\Gamma}_2$ and not to the entire shape image $I$.

Eventually, via application of the introduced method, the optimal fit of curves $\Gamma_1$ and $\Gamma_2$ is achieved, where curvature similarity, rotation, $x$ and $y$ parallel translation altogether have been taken into account. A number of results of such a fitting procedure in association with contours of letters appearing on ancient inscriptions and Byzantine codices are displayed in figures 3,4.

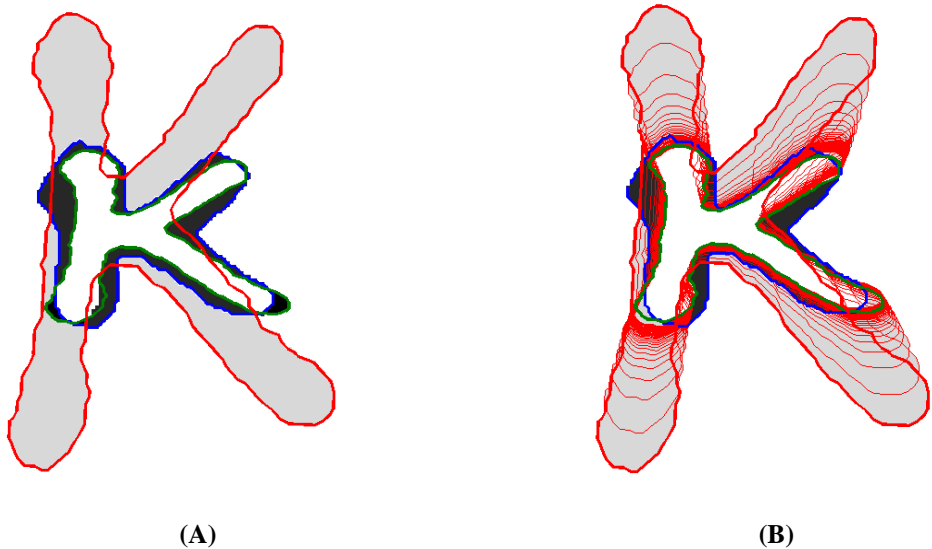

(A)  (B)

**Fig. 3** Depiction of the matching procedure applied to two letter contours belonging to different codices.

**A)** Initial and optimal placement of two "kappa" symbol contours. The reference, fixed contour is depicted in blue. The initial position of the curve to be matched is shown in red and its optimal transformed version in green.

**B)** Intermediate deformations of the transformed contour towards convergence, are presented by the curves in red. The final curve of this deformation process is rotated and translated, so as to optimally fit the reference one.

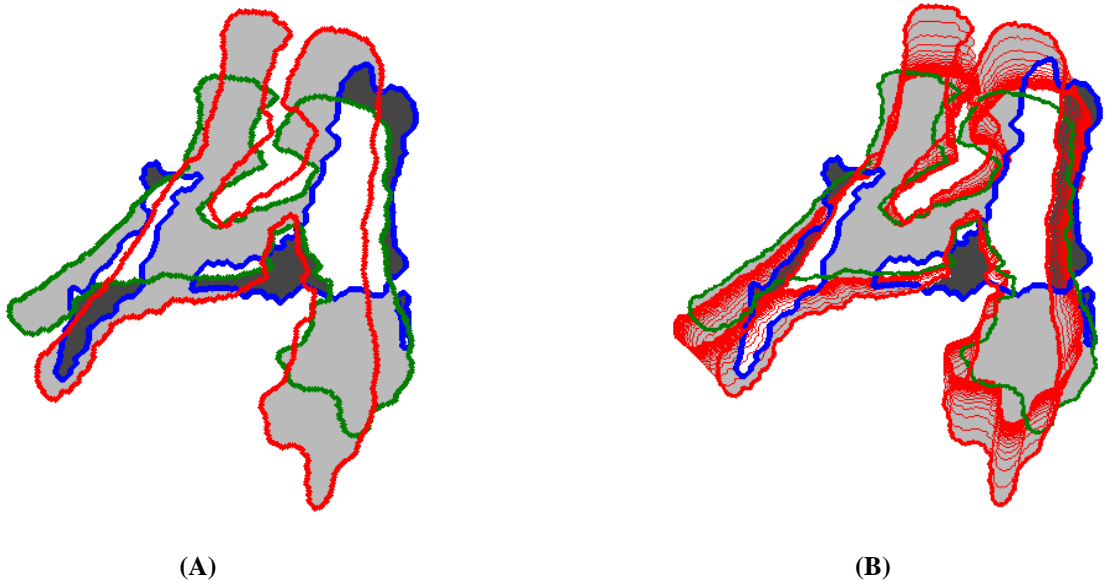

<div align="center">

**(A)**             **(B)**

</div>

**Fig. 4** Depiction of the matching procedure applied to two letter contours belonging to different inscriptions.

**A)** Initial and optimal placement of two "alpha" symbol contours. The reference, fixed contour is depicted in blue. The initial position of the curve to be matched is shown in red and its optimal transformed version in green.

**B)** Intermediate deformations of the transformed contour towards convergence, are presented by the curves in red. The final curve of this deformation process is rotated and translated, so as to optimally fit the reference one.

## 5.4 The introduced measures of similarity of two curves.

After obtaining the optimal position of $\widetilde{\Gamma}_2$ and $\Gamma_1$, if one employs the previously defined fitting errors as similarity measures of curves $\Gamma_1$ and $\Gamma_2$, then one will meet with serious inconsistencies; we stress that these inconsistencies are associated with the measure of similarity of curves $\Gamma_1$ and $\Gamma_2$ and not with the best fit of these curves, which is optimally achieved by means of the method introduced in 5.2 and 5.3. In fact, the various alphabet symbols realizations inside the very same manuscript manifest substantial variation in size and curvature distribution of the contour. Therefore, if one matches all realizations of the same alphabet symbol in a manuscript to a specific realization contour $\Gamma_1$, then the obtained errors will immediately depend on the size and mean curvature of the reference letter. In such a case, statistical processing of the obtained error values may be meaningless. Equivalently, in order to apply a consistent statistical decision, like the one that will be introduced in the following Section, one must first define a measure of similarity of $\Gamma_2$ and $\Gamma_1$ that is as much independent as possible of the size and mean curvature of the chosen reference curve $\Gamma_1$. For this reason, we have defined and employed the following measures of similarity of $\widetilde{\Gamma}_2$ and $\Gamma_1$:

**DEFINITION 4:**

*a)     Similarity measure in plane curvature space*

Let $|\zeta^C|_M$ be the obtained minimum value of plane curvature difference between $\widetilde{\Gamma}_2$ and $\Gamma_1$ and consider the already computed $\varepsilon_1 = \oint_{\Gamma_1} \nabla \cdot \vec{n} \; dl$, namely the integral of the plane curvature on the nearest to $\Gamma_1$ isocontour, towards $\Gamma_2$, in the digital domain $I$; this integral has already been computed in Step 2 of Section 5.2.

Then, we define the similarity measure of $\widetilde{\Gamma}_2$ and $\Gamma_1$, as far as plane curvature difference is concerned, via relation: $|\zeta^C|_S = \dfrac{|\zeta^C|_M}{\varepsilon_1}$.

*b)     Similarity measure in Euclidean space*

Consider $\widetilde{\Gamma}_2$ rotated and translated, so as to optimally fit $\Gamma_1$ according to the previously defined Euclidean error. If $\left( \oint_{\widetilde{\Gamma}_2} d(P)dl \right)_M$ is the corresponding minimum fitting error, then the similarity measure of $\widetilde{\Gamma}_2$ and $\Gamma_1$ is defined via: $\theta_M(\Gamma_1, \widetilde{\Gamma}_2) = \dfrac{\left( \oint_{\widetilde{\Gamma}_2} d(P)dl \right)_M}{\sqrt{\ell(\Gamma_1)\ell(\widetilde{\Gamma}_2)}}$

Where $\ell(\Gamma_1)$ and $\ell(\widetilde{\Gamma}_2)$ are the lengths of curves $\Gamma_1$ and $\widetilde{\Gamma}_2$ respectively.

**DEFINITION 5:** *The overall similarity measure of $\Gamma_1$ and $\widetilde{\Gamma}_2$*

The overall similarity measure of the two curves $\Gamma_1$ and $\widetilde{\Gamma}_2$ is given by:

$$\xi(\Gamma_1, \widetilde{\Gamma}_2) = |\zeta^C|_S + \rho \, \theta_M(\Gamma_1, \widetilde{\Gamma}_2),$$

Where $\rho$ is a constant, properly chosen to ensure that the two errors $|\zeta^C|_S$ and $\theta_M(\Gamma_1, \widetilde{\Gamma}_2)$ are of similar order of magnitude.

We will plausibly adopt that $\xi(\Gamma_1, \widetilde{\Gamma}_2)$ is an implicit measure of the similarity of the two initial curves $\Gamma_1$ and $\Gamma_2$; this claim is fully supported by the analysis and the results introduced in Sect. 5.1. Actually $\theta_M(\Gamma_1, \widetilde{\Gamma}_2)$ is the geodesic path error, when $F(x, y)$ is the Euclidean distance transform, while $|\zeta^C|_S$ is the supremum of the first order variation of this error. The efficiency of the chosen similarity measure will also be justified by the final writer identification results.

# 6.    Statistical criteria for classifying a given set of shapes into groups – Application to writer identification

## 6.1    Determination of the different writers of a given set of documents

Let us consider two distinct documents $D_1$ and $D_2$ and a specific alphabet symbol, say $L$, $N_1^L$ realizations of which appear in $D_1$, while $N_2^L$ realizations of it appear in $D_2$.

We arbitrarily choose a first realization of $L$ in $D_1$, we call it $L_{1,1}$ and we perform all pair-wise comparisons of it with all other realizations of $L$ on $D_1$ namely $L_{1,2}, L_{1,3}, \ldots, L_{1,N_1^L}$. In order to perform these comparisons, we embed $L_{1,1}$ and $L_{1,i}$ in the same domain $I$, which may be thought of as a rectangular white digital image. We apply to this couple of curves the method introduced in Sections 2, 3, 4, 5 and, thus, we obtain a similarity measure of the type described in Section 5.4, which we denote $\xi(L_{1,1}, \widetilde{L}_{1,i})$. Subsequently, we consider realization $L_{1,2}$ as the reference one and we compare it with $L_{1,3}, L_{1,4}, \ldots, L_{1,N_1^L}$, by means of the introduced method, thus, obtaining a set of similarity measures $\xi(L_{1,2}, \widetilde{L}_{1,i})$, $i > 2$ and so forth, until $L_{1,N_1^L-1}$ is compared with $L_{1,N_1^L}$. Associated optimal fitting positions of pairs of contours of various alphabet symbols realizations are shown in figures 3a, 4a.

We assume that the obtained $\frac{(N_1^L - 1)N_1^L}{2}$ similarity measures $\xi(L_{1,j}, \tilde{L}_{1,i})$, $i > j$ come from a normal distribution; the performed Kolmogorov-Smirnov test did not violate this assumption ($a$=0.01) as well as the related simulation experiments (see figure 5). We let $\xi\mu_1$ be their mean value of $\xi(L_{1,j}, \tilde{L}_{1,i})$ and $\xi S_1$ be their standard deviation.

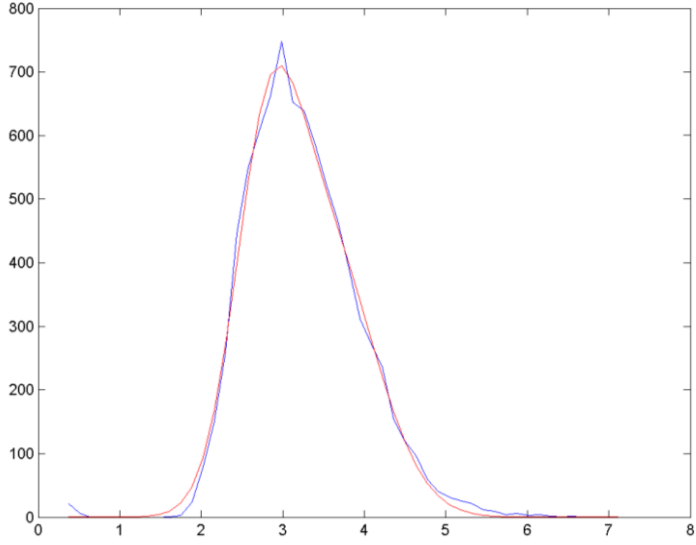

**Fig. 5** : Histogram of the similarity measure, shown in blue, together with the theoretical normal distribution optimally fit to it, shown in red.

As a next step, we apply the method to all pairs of realizations $L_{1,i}$, $i = 1,2,...,N_1^L$ and $L_{2,j}$, $j = 1,2,...,N_2^L$ and we obtain another set of values $\xi(L_{1,i}, \tilde{L}_{2,j})$, $i = 1,2,...,N_1^L$, $j = 1,2,...,N_2^L$. We again assume that the obtained $N_1^L N_2^L$ values $\xi(L_{1,i}, \tilde{L}_{2,j})$ come from a normal distribution; the performed Kolmogorov-Smirnov test did not violate this assumption ($a$=0.01) as well as the related simulation experiments. We let $\xi\mu_2$ be their mean value of $\xi(L_{1,i}, \tilde{L}_{2,j})$ and $\xi S_2$ be their standard deviation.

Then, if $\mu_1$, $\mu_2$ are the (unknown) theoretical means of the two normal populations and if $N_{1,1}^L = \frac{(N_1^L - 1)N_1^L}{2}$ and $N_{1,2}^L = N_1^L N_2^L$, it is well known that quantity

$$t^L = \frac{(\xi\mu_1 - \xi\mu_2) - (\mu_1 - \mu_2)}{\sqrt{\dfrac{(\xi S_1)^2}{N_{1,1}^L} + \dfrac{(\xi S_2)^2}{N_{1,2}^L}}} \tag{6.1}$$

follows a Student distribution with $d$ degrees of freedom, where $d$ is the integral part of

$$\frac{\left(\dfrac{(\xi S_1)^2}{N_{1,1}^L} + \dfrac{(\xi S_2)^2}{N_{1,2}^L}\right)^2}{\dfrac{\left(\dfrac{(\xi S_1)^2}{N_{1,1}^L}\right)^2}{N_{1,1}^L - 1} + \dfrac{\left(\dfrac{(\xi S_2)^2}{N_{1,2}^L}\right)^2}{N_{1,2}^L - 1}} \tag{6.2}$$

As it has already been noted, the population theoretical means $\mu_1$, $\mu_2$ are, as a rule, not known; nevertheless, if one makes the hypothesis that the two documents $D_1$ and $D_2$ have been written by the same hand, then, a priori, $\mu_1 = \mu_2$. In this way, quantity $t^L$ in (6.1) has a well-defined value and therefore, the validity of the hypothesis $H_0$, where $H_0 : \mu_1 = \mu_2$, can be tested against the complementary hypothesis $H_1 : \mu_1 \neq \mu_2$.

The degree of confidence with which these hypotheses will be tested is chosen by heuristic arguments. Indeed, in practice, since inscribing stones was a profession up to the Hellenistic Period, one may safely consider that the ensemble of all unearthed inscriptions so far, were inscribed by a maximum of few hundreds of writers. At the same time, one may expect that a quite complete database of these inscriptions should include some tenths of thousands of them. Thus, since multiple comparisons will be performed, as it will be described below, if we also take the Bonferoni approach [40] into consideration, a logical value for the threshold of the level of significance $a$ is $a_T = \dfrac{10^{-4}}{\text{n}}$, where 'n' is the number of the performed statistical tests associated with the number of different alphabet symbols, on which the introduced methodology is each time applied. Given that analogous arguments hold for the Byzantine codices too, the same $a_T$ value can be used as a threshold, also for the considered codices. Therefore, if

$P(-\infty < x < -|t^L| \vee |t^L| < x < +\infty) < a_T$, then $H_0$ is rejected and $H_1$ is adopted; otherwise, one cannot reject $H_0$, with a considerable confidence.

Next, we proceed as follows in order to take into account the entire set of $\nu$ considered documents $\Sigma_0 = \{D_1, D_2, ..., D_\nu\}$: we apply the introduced methodology to each pair of documents $(D_1, D_p)$, $p = 2, ..., \nu$ for all alphabet symbols $L$ appearing in the entire set of documents.

We repeat the same procedure, employing now document $D_2$, as a reference one. Then, $D_3$, subsequently $D_4$ and so forth, until document $D_{\nu-1}$ is compared with $D_\nu$ for all selected, common to all documents, alphabet symbols $L$.

For every pair of documents, for which $H_0$ is rejected and for each comparison between L appearing in them, we keep the one with minimum $P\left(-\infty < x < -\left|t^L\right| \vee \left|t^L\right| < x < +\infty\right)$, provided that this probability is smaller than $a_T$; let this pair of inscriptions be $\left(D_{q_1}, D_{q_2}\right)$. Then, reasonably enough, we assume that $D_{q_1}$ and $D_{q_2}$ have been written by two different hands, say $W_1$ and $W_2$. We let document $D_{q_1}$ be the first document associated with writer $W_1$ and actually, we assume that this document is a first good representative of its writing style and for this reason, we denote $D_{q_1}$ with the alternative symbol $\Delta_1$. Similarly, we let $D_{q2}$ be the first good representative of the writing style of $W_2$ and for this reason, we denote it with the alternative symbol $\Delta_2$.

Subsequently, we remove documents $D_{q_1}$ and $D_{q_2}$ from the ensemble of the considered documents $\Sigma_0$, thus, reducing the ensemble of the manuscripts to-be-classified to $\Sigma_1 = \Sigma_0 - \{D_{q_1}, D_{q_2}\}$. At this point, we take into consideration all pair-wise comparisons of each unclassified document of set $\Sigma_1$ with manuscripts $\Delta_1$ and $\Delta_2$, equivalently with documents $D_{q_1}$, $D_{q_2}$. In fact, consider an arbitrary $D_i \in \Sigma_1$ and let it be matched with $\Delta_1$ first and $\Delta_2$ next, by means of the introduced methodology. In this way, for each alphabet symbol $L$ separately, two similarity measures, two corresponding values $t_{i,1}^L$, $t_{i,2}^L$ (via 6.1) and two corresponding tail probabilities $P_{i,1}\left(-\infty < x < -\left|t_{i,1}^L\right| \vee \left|t_{i,1}^L\right| < x < +\infty\right)$ and $P_{i,2}\left(-\infty < x < -\left|t_{i,2}^L\right| \vee \left|t_{i,2}^L\right| < x < +\infty\right)$ are obtained. We choose the maximum of $P_{i,1}$ and $P_{i,2}$ and then the minimum of all these maxima for all alphabet symbols L appearing in the compared documents, i.e. $\alpha_i = \min_{over\ all\ L}\{\max\{P_{i,1}, P_{i,2}\}\}$.

Now, we consider all $\alpha_i$ with value smaller than the threshold of the level of significance $\alpha_T$, i.e. all $\alpha_i \leq \alpha_T$ and we choose the minimum of these $\alpha_i$ that corresponds, say, to manuscript $D_{q_3}$. Then, we assume that $D_{q_3}$ belongs to a third writer, denoted by $W_3$, adopting that $D_{q_3}$ is a good representative of the writing style of $W_3$; for this reason, we re-symbolize $D_{q_3}$ as $\Delta_3$. In other words, in case that there is at least one alphabet symbol in $D_{q_3}$, whose realizations are drastically different from the realizations of this symbol on both $\Delta_1$ and $\Delta_2$, then we decide that document $D_{q_3}$ belongs to a different writer, with a considerable confidence.

Proceeding in an analogous manner, we remove document $D_{q_3}$ from $\Sigma_1$, thus obtaining the set $\Sigma_2 = \Sigma_1 - \{D_{q_3}\} = \Sigma_0 - \{D_{q_1}, D_{q_2}, D_{q_3}\}$. We take into consideration the results of all pair-wise comparisons between each document of $\Sigma_2$ with $\Delta_1, \Delta_2, \Delta_3$, in connection with all alphabet symbols L appearing in the compared documents. In this way, we obtain

$$\alpha_i = \min_{over\ all\ L}\{\max\{P_{i,1}, P_{i,2}, P_{i,3}\}\}$$

and we spot the minimum of $\alpha_i$, provided that $\alpha_i \leq \alpha_T$, if any. Consequently, we spot a new writer $W_4$, with $\Delta_4$ being a good representative document of his writing style.

We continue the process until inequality $\alpha_i \leq \alpha_T$ is not satisfied for any comparison whatsoever. In this case, we end up with a number of writers (different hands) $W_1, W_2, ..., W_\lambda$ and their corresponding representative documents $\Delta_1, \Delta_2, ..., \Delta_\lambda$. We would like to stress that, on the considered application, the determination of different hands is quite independent of the choice of the $\alpha_T$ value, the only restriction being that the order of magnitude of $\alpha_T$ is reasonable. In other words, when the number of distinct hands has been reached, then an important "jump" in the order of $\alpha_i$ is observed; thus, the group of the representative documents $\Delta_1, \Delta_2, ..., \Delta_\lambda$ can quite safely be distinguished.

In practice, at the end of this process, there will be a set of unclassified documents $\Sigma_U = \Sigma_0 - \{D_{q_1}, D_{q_2}, ..., D_{q_\lambda}\}$; therefore, it is necessary to develop a method for classifying the documents of $\Sigma_U$ to the proper writer $W_1$ or $W_2$ or…or $W_\lambda$.

### 6.2 Classification of the remaining documents to the proper writer

Let us consider any document $D_i \in \Sigma_U$; we compare $D_i$ with each one of $\Delta_1, \Delta_2, ..., \Delta_\lambda$ by means of the introduced method. More specifically, consider $D_i$ and document $\Delta_1$ and suppose that there are realizations of alphabet symbol 'A' appearing on both these documents. Then, we momentarily assume that $D_i$ and $\Delta_1$ have been written by the same hand, in which case, the theoretical population means $\mu_1$ and $\mu_2$ in (6.1) are equal. Under this assumption, the value of $t_{i,1}^A$ is known and, consequently, the value of the underlying Student distribution at this point, $f^{St}\left(t_{i,1}^A\right)$ is known. We repeat this procedure for alphabet symbols appearing both in $D_i$ and $\Delta_1$; let $\rho_{i,1}$ be the number of these common alphabet symbols denoted by $G_1, G_2, G_{\rho_{i,1}}$; in the previous examples $G_1 = 'A'$, $G_2 = '\Omega'$, etc. Then, we define the likelihood of similarity of documents $D_i$ and $\Delta_1$, as far as the writing style is concerned, by means of

$$\vartheta_{i,1} = \sqrt[\rho_{i,1}]{\prod_{p=1}^{\rho_{i,1}} \left(f^{St}(t_{i,1}^{Gp})\right)} \qquad (6.3)$$

This is so, because if $D_i$ and $\Delta_1$ have been written by the same hand, then it is unlikely that $t_{i,1}^{Gp}$ will be far away from the $y$-axis. On the contrary, it is logical to assume that if they have not been written by the same hand, then $t_{i,1}^{Gp}$ would have a particularly small value for at least a number of common alphabet symbols $G_p$. We employ the root of order $\rho_{i,1}$ to ensure that the order of the likelihood $\vartheta_{i,1}$ is independent of the number of the common symbols between $D_i$ and $\Delta_1$. We repeat this process for all pairs of compared documents $D_i$ and $\Delta_j$, thus obtaining a corresponding likelihood value $\vartheta_{i,j}$. We compute the maximum of $\vartheta_{i,j}$ overall $j$ and say that this maximum occurs at $j = q$; namely, comparison of document $D_i$ with representative document $\Delta_q$ offered the maximum likelihood $\vartheta_{i,q}$. In this way, we attribute document $D_i$ to writer $W_q$ who has $\Delta_q$ as his representative document. Next, we remove $D_i$ from $\Sigma_U$ and we repeat the aforementioned process until no more unclassified documents remain.

### 6.3 Classifying a given set of curves according to an arbitrary property $\mathcal{P}$.

The previous analysis presented in sub-Sections 6.1 and 6.2 refers to the classification of alphabet symbols according to their writer. However, this approach may be directly generalized, so as to tackle the problem of classifying an ensemble of arbitrary shapes according to a property $\mathcal{P}$, having the following characteristics: property $\mathcal{P}$ divides the ensemble of given curves into distinct families; if a member of each such family is subdue to rotation, parallel translation or resize along the $x$ and $y$-axis separately, then the corresponding curve remains inside the very same family.

The introduced method does not require any database of reference shapes or any a priori knowledge, except of a plausible choice of the threshold $a_T$ of the level of significance. In other words, given an ensemble of curves, which do not necessarily represent letters, in order to optimally match any two of them, we first apply the plane curvature methodology introduced in Sections 2, 3, 4 and 5.2, thus obtaining the optimal resize factors $a$ and $b$.

Afterwards, we apply rotation and parallel translation to the already resized curve and we optimally place the two compared curves, so as the corresponding error, as described in Section 5, is minimum. Then, we spot the statistical distribution of similarity measure $\xi$; if property $\mathcal{P}$ is multi-parametric, then we assume that $\xi$ follows a normal distribution due to the Central Limit Theorem. In any case, one may test the theoretical assumption about the statistical distribution by employing the Kolmogorov-Smirnoff test.

As a next step, we apply proper statistical tests to determine the number of distinct families of curves according to property $\mathcal{P}$. At this point, selection of a plausible value for the threshold $\alpha_T$ of the level of significance is crucial and depends on a certain "high level" knowledge of property $\mathcal{P}$. This process will attribute to each family of curves a representative curve corresponding to the minimum two-tailed area of the performed statistical comparisons or equivalently the curve that differs most from the others, statistically.

Finally, unclassified curves are attributed to the proper family by means of a maximum likelihood approach employing a likelihood measure of the type of (6.3); evidently, the underlying probability density functions will correspond to characteristics of property $\mathcal{P}$, but they will always be those of the type of (6.1) with degrees of freedom analogous to (6.2), when $\xi$ follows a normal distribution.

## 6.4 Time complexity of the whole curves' classification system

Concerning the time complexity of the classification methodology, described in Sects. 6.1 and 6.2, we should note that the whole system mainly splits into 2 sub-procedures : A) First, optimal affine registration is performed for all pairs of curves and the fitting error of each pair is computed and stored. B) Next, using all pair-wise fitting error results, the probabilistic classification procedure of Sects. 6.1 and 6.2 sequentially classifies the considered curves into groups. Evidently, the first procedure is parallelizable by simple distribution of the data in a 2D processors grid and requires minimal message passing (i.e. just to read the distributed data and to publish the results when execution ends). Concerning the classification procedure, we note that it is sequential and thus, it cannot be parallelized. However, the likelihood calculations are performed pretty rapidly and the corresponding execution time is by no means comparable to the execution time of the first procedure. Actually, implementing the methodology using C in a computer with a processor of 2 cores, each clocked at 2.8 GHz and with 6MB L2 cache, the first procedure returned all pair wise comparisons of the instances of an alphabet symbol in any 2 documents in less than 10 sec. Moreover, execution time of the statistical classification procedure in both writer identification applications did not exceed 0.5 sec. Evidently, the smaller the documents, the faster the execution time; in certain cases, this procedure lasted for less than a second.

We once more state that we have applied the introduced methodology in 46 inscriptions and 23 byzantine codices. The overall number of tested alphabet symbols realizations was greater than 10000 for the inscriptions and more than 5000 for the byzantine codices. We would like to stress, however, that the average number of frequently encountered letters, such as A, E, Ω, etc. per document was around 30, while there were letters appearing 2-3 times per document. Therefore, the number of associated comparisons varies in an analogous manner.

The resolution of the letters' images was around 260 pixels per cm for the inscriptions and around 160 pixels per cm for the byzantine codices, while the images' dimensions were vastly around 300x300 pixels and 90x90 pixels correspondingly.

# 7.    Application of the introduced methodology to public reference datasets

In this Section, the performance of the introduced methodology for the classification of 2D shapes is tested in 2 large data sets. First, the MNIST database of digits [41] is considered. Since the introduced methodology is developed so as to overcome absence of training set, the classification of the MNIST test set images is performed on the basis of their comparison with a reduced random collection of images from database's training set. In terms of the description of the methodology given in Sect. 6, the randomly selected test set instances of each digit are treated as instances of the same letter from different documents (i.e. digits of the same type are treated as documents of the same writer). We should emphasize here that the problem of recognizing the same character for all different writers is essentially different from the problem of recognizing the writer of the different instances of the same character. Namely, variation of the instances of a symbol written by the same writer is expected to be non causal, i.e. it is noise, when affine transformations are normalized. On the other hand, shapes of the instances of a character written by different writers are expected to exhibit conceptual differences even if affine transformations are optimally normalized. Thus, the classification results we will present, mainly evaluate the efficiency of the affine registration procedure and not of the probabilistic classification scheme, since the digits grouping is performed as independent single character documents' classification. So, each digit of the test set is compared with the elements of all digits' groups, randomly formed from the Training Set. Then, any considered test digit is attributed to the group whose elements exhibit the least mean fitting error. The classification procedure has been repeated for many different selections of the digits groups' elements so as to suppress randomness in the classification results. The average success rate of these classifications is kept to evaluate the performance of the introduced methodology, under the previously stated restrictions. Since, we are using a limited part of the training set each time, the classification's performance depends on the size of the employed training set part. In order to evaluate this relation, we have repeated the classification task with training set groups of ascending size. The performance of the Test Set digits classification for 5 different Training Set's groups of growing size is summarized in Table 1.

*TABLE 1: MNIST Test Set Classification Results*

| Number of Training Set's groups' elements | Test Set's classifications' average success rate (%) | Standard Deviation of classifications' average success rate (%) |
|---|---|---|
| 10 | 80.11 | 1.64 |
| 20 | 85.28 | 0.93 |
| 30 | 93.41 | 0.61 |
| 40 | 98.61 | 0.34 |
| 50 | 99.49 | 0.11 |

The second considered dataset of shapes is the MPEG-7 Core Experiment CE-Shape-1. It consists of 1400 shapes belonging to 70 groups of equal size. As in the case of the digits recognition problem, the shapes that MPEG7 groups consist of, also exhibit non rigid deformations. Hence, the statistical classification scheme of Sect. 6 is not applicable to the grouping of dataset's shapes, since, even if affine transformations are normalized, the variations of the shapes of the same group are causal non rigid deformations and thus they are not elements of a statistical distribution. So, using the affine registration scheme introduced in Sects. 5.2, 5.3 and the shapes similarity measures defined in Sect. 5.4, classification of the dataset's shapes into groups is performed on the basis of the average fitting error results. Namely, for a specific shape of the dataset, all pair wise matches of it with the other shapes are considered. Then, this shape is classified into the group whose elements record the best average fitting error results. The classification's success rate was 87%.

## 8.    Application of the introduced methodology to the identification of the writer of ancient inscriptions and byzantine codices

In this Section, we will employ the previously obtained results in identifying the writers of a set of important ancient inscriptions and an ensemble of equally important Byzantine codices. We will first give a brief description of the selected sets of ancient documents, together with information concerning their selection.

In particular, Prof. Stephen Tracy, ex-Director of the American School of Classical Studies at Athens, now professor at the School of Historical Studies of Institute for Advanced Study at Princeton, USA, has chosen a number of ancient inscriptions, while Prof. Christopher Blackwell, Professor of Classics at Furman University in Greenville, South Carolina, has selected a number of Byzantine codices. We would like to emphasize that both Prof. St. Tracy and Prof C. Blackwell were very rigorous, severe and very careful at not disclosing any information whatsoever to the rest of the authors, concerning the documents upon which the methodology has been tested; for example, they did not reveal anything about the era of each document, its content, the place where it was found or any other information they already knew. Indeed, the entire information about the documents exposed in sub-section 7.1 below, has been disclosed and written by these two authors, after the application of the method and the presentation of the related results by the rest of the team.

## 8.1 The selected set of documents

A list of the selected inscriptions is shown in Table 2, where each inscription is labeled by its code number in Agora and the Epigraphic Museum of Greece (Original ID). The 46 inscriptions chosen by Professor Tracy represent a fairly broad time period, namely from mid-fourth century B.C. to late second century B.C. and a variety of different styles of lettering.

*TABLE 2: List of the inscriptions sorted by number*

| Work. ID | Orig. ID | Work. ID | Orig. ID | Work. ID | Orig. ID | Work. ID | Orig. ID | Work. ID | Orig. ID |
|---|---|---|---|---|---|---|---|---|---|
| I1 | 0247 | I11 | 4330 | I21 | 6422 | I31 | 7335 | I41 | 7542 |
| I2 | 0286 | I12 | 4424 | I22 | 6671 | I32 | 7398 | I42 | 7566 |
| I3 | 1024 | I13 | 4462 | I23 | 7041 | I33 | 7400 | I43 | 7567 |
| I4 | 1640 | I14 | 4917 | I24 | 7156 | I34 | 7405 | I44 | 7587 |
| I5 | 2054 | I15 | 5039 | I25 | 7188 | I35 | 7446 | I45 | 7723 |
| I6 | 2361 | I16 | 5297 | I26 | 7190 | I36 | 7457 | I46 | 10068 |
| I7 | 3717 | I17 | 6006 | I27 | 7220 | I37 | 7478 | | |
| I8 | 3855 | I18 | 6053 | I28 | 7237 | I38 | 7481 | | |
| I9 | 4033 | I19 | 6124 | I29 | 7245 | I39 | 7482 | | |

| I10 | 4266 | I20 | 6295 | I30 | 7254 | I40 | 7519 | |

A list of the selected Byzantine codices is shown in Table 3, where each manuscript is labeled by its code number given by the Technical Institution it comes from (Original ID)

*TABLE 3: List of the byzantine codices sorted by number*

| Working ID | Original ID | Working ID | Original ID | Working ID | Original ID | Working ID | Original ID |
|---|---|---|---|---|---|---|---|
| **BC1** | T01_I01 | **BC7** | T02_I08 | **BC13** | T03_I17 | **BC19** | T04_I23 |
| **BC2** | T01_I02 | **BC8** | T02_I09 | **BC14** | T03_I18 | **BC20** | T04_I24 |
| **BC3** | T01_I04 | **BC9** | T02_I10 | **BC15** | T03_I19 | **BC21** | T04_I25 |
| **BC4** | T01_I05 | **BC10** | T02_I11 | **BC16** | T03_I20 | **BC22** | T04_I26 |
| **BC5** | T01_I06 | **BC11** | T02_I13 | **BC17** | T04_I21 | **BC23** | T04_I27 |
| **BC6** | T01_I07 | **BC12** | T03_I14 | **BC18** | T04_I22 | | |

## 8.2 Intrinsic difficulties in the identification of writers of ancient documents

One major difficulty faced by the authors was that there is no training set available, namely no manuscripts at all were used as reference, in contrast with most approaches appearing in the bibliography so far (see Section 1). Equivalently, no text written by the sought-for hands was used as a reference document and no pre-existing database of information concerning the texts or their writers has been employed.

Another major difficulty faced by the authors was that they had no previous knowledge of the number of distinct hands who had written the considered manuscripts.

In addition, some of the most important problems emerging when trying to automatically identify the writer of ancient documents are:

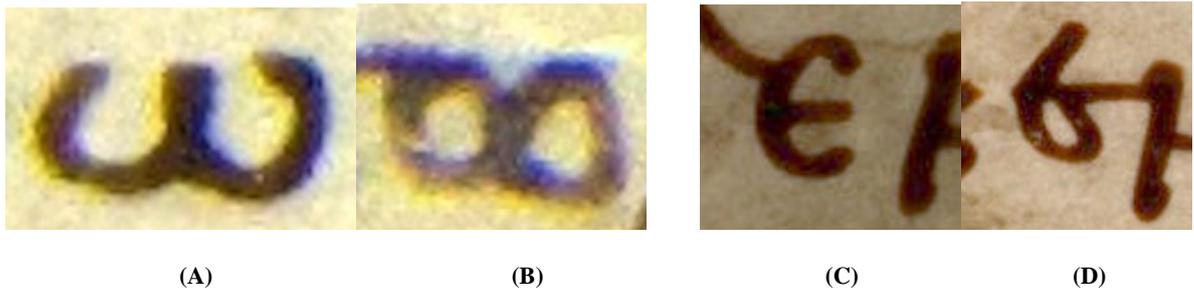

| (A) | (B) | (C) | (D) |

**Fig. 6** : Manifestation of shape variability encountered among realizations of the same alphabet symbol generated by the same writer.

**(1)** The obvious variability of the way a writer forms an alphabet symbol; for example, the realizations of letter $\Theta$ or $\Omega$ or $E$, even if they are written by the same hand and even if they are encountered in the same manuscript, can be either closed or open (see figure 6).

Moreover, in a number of documents, alphabet symbol kappa looks like the Latin *'u'*, while somewhere else in the same document, it looks like $\kappa$. Similarly, in numerous ancient inscriptions, as far as letter $A$, the same writer connects the middle cross bar of $A$ a number of times with the left leg only, other times with the right leg only, while sometimes he does not connect it with any leg at all, etc.

**(2)** The contours of the realizations of an alphabet symbol, especially in inscriptions but also in certain byzantine codices, are, as a rule, noisy (see Figure 7). This noise may be due to many factors, such as the quality of the employed stone or writing material, the precise form of the writing instrument (chisel, pen, etc), the age of the writer, his fatigue and mood, etc.

**(3)** It is frequently observed that the similarity between two specific samples of an alphabet symbol of the same writer is smaller than the similarity between other pairs of samples by different writers.

**(4)** The considered manuscripts may suffer from serious wear (see Figures 8).

**(5)** Based on the aforementioned remarks, one must determine a kernel in each alphabet symbol realization that remains invariant in the different documents written by the same writer. Moreover, a mathematical description of this kernel must be found and expressed.

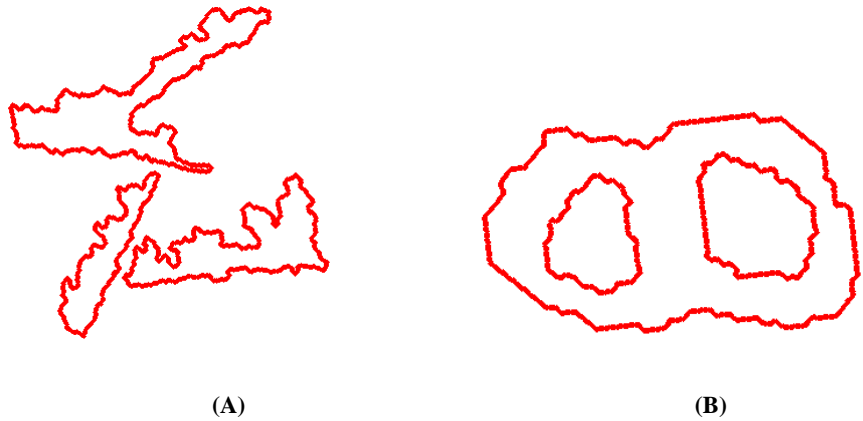

<div align="center">

**(A)**            **(B)**

</div>

**Fig. 7** : Manifestation of noisy contours encountered among the realizations of alphabet symbols in both inscriptions and Byzantine codices.

### 8.3 The identification results obtained via application of the introduced method

First, we would like to point out that for the application in hand, the contour and the body of the various alphabet symbols realizations are considered given. However, we note that all images of the considered alphabet symbols realizations have been segmented by the method introduced in [42]. In addition, the black and white versions of the letter images, as well as the corresponding letters' contours have been obtained by the method presented in Section 2 of [43].

The application of the introduced methodology upon both ancient inscriptions and Byzantine Codices, offered a number of really successful results, as far as writer identification is concerned. In particular, in connection with the considered inscriptions, the exhaustive pair-wise Bonferoni-type comparisons, as described in Section 6, led us to the conclusion that all 46 inscriptions have been written by ten (10) distinct writers, as described in table 4.

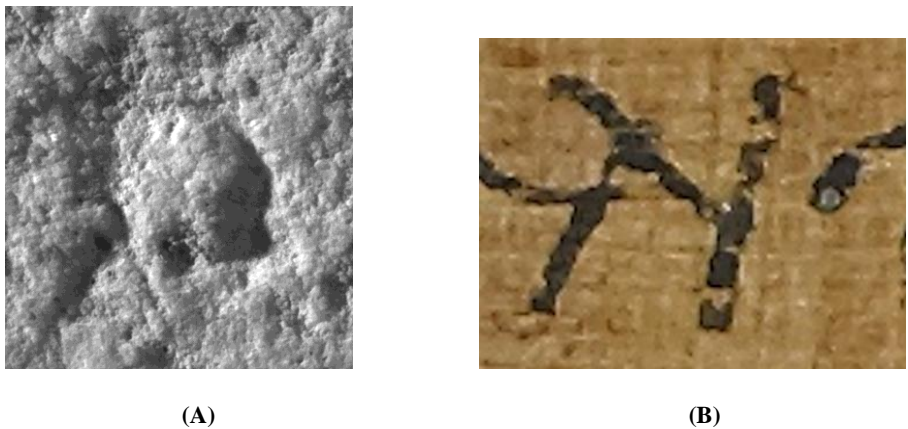

<div align="center">

**(A)**                    **(B)**

</div>

**Fig. 8** : Manifestation of the wear encountered in the realizations of alphabet symbols in both inscriptions and Byzantine codices.

*TABLE 4. The final classification of inscriptions according to their writer*

| Individual writers | Inscriptions classified to each writer |
|---|---|
| Hand 1 | I33, I34, I41, I19, I32 |
| Hand 2 | I9, I38, I39, I46, I6, I4 |
| Hand 3 | I17, I20, I40, I42, I43, I16, I18 |
| Hand 4 | I28, I31, I30 |
| Hand 5 | I23, I26, I27 |
| Hand 6 | I36, I35, I29, I45 |
| Hand 7 | I1, I22, I14, I25, I13, I7 |
| Hand 8 | I10, I12, I44, I15, I11 |
| Hand 9 | I2, I21, I24, I37 |
| Hand 10 | I3, I8, I5 |

Prof. Tracy fully agrees with this classification, with no exception whatsoever, as well as other prominent epigraphists. We would like to stress that a scholar usually needs years of work to achieve such a classification.

Analogous results have been obtained in association with the examined Byzantine codices. More specifically, after the application of the introduced methodology, as it was described in Sections 6.1 and 6.2, the conclusion has been reached that four (4) distinct hands have written all 23 codices; the corresponding representative Byzantine manuscripts for each one of these hands are $\widetilde{D}_1 \overset{\text{def}}{=} \text{BC}_{13}$, $\widetilde{D}_2 \overset{\text{def}}{=} \text{BC}_{19}$, $\widetilde{D}_3 \overset{\text{def}}{=} \text{BC}_{23}$ and $\widetilde{D}_4 \overset{\text{def}}{=} \text{BC}_3$. The remaining 19 Byzantine codices have been classified to one of these 4 hands by the maximum likelihood criterion introduced in section 6.3, as described in table 5.

Prof. Christopher Blackwell and experts of the Center of Hellenic Studies (CHS) of Harvard University fully agreed with this classification.

*TABLE 5. The final classification of byzantine codices according to their writer*

| Individual writers | Inscriptions classified to each writer |
|---|---|
| Hand 1 | BC1, BC5, BC8, BC11, BC12, BC18 |
| Hand 2 | BC2, BC3, BC6, BC7, BC13 |
| Hand 3 | BC4, BC9, BC13, BC15, BC22, BC23 |
| Hand 4 | BC14, BC17, BC19, BC20 |

**8.4    Comparison of the efficiency of the introduced similarity measures with other shapes similarity metrics.**

In addition to the efficiency test of the introduced similarity measures in the case of shapes retrieval, in this section we test how well efficient methodologies of shapes retrieval and classification behave in the case of the considered inscriptions – byzantine codices. Specifically we have chosen the method of "Inner Distance shape context" (IDSC) [44] and the method of [45] for improving the classification results offered by the IDSC. We note that the implementation used for IDSC is the one distributed by the authors in http://www.dabi.temple.edu/~hbling/code_data.htm, while the implementation used for the method of [45] is the one distributed in http://www.loni.ucla.edu/~ztu/. Similarity measurements have been performed as described in Sect. 6.1, i.e. by exhaustive pair-wise comparisons between different documents for each considered alphabet symbol separately. Then, the efficiency of the results has been tested against the ground truth of the classification obtained by the introduced methodology and verified by the expert epigraphists.

In particular, we have calculated the mean values of the similarity measurements for all pairwise comparisons between the hands detected both in the inscriptions and in the byzantine codices dataset. Next, the efficiency of the similarity measures is evaluated by testing the minimization of the related error metrics at the correct groups. These results for the data set of byzantine codices and for letters "kappa", "omega", "epsilon", indicatively selected, are presented in tables 6.1-6.3 column-wise (i.e. the similarity measurements of a hand with all others are presented as column vectors). If dissimilarity is minimized (or similarity is maximized) when testing similarity of the documents of the same hand, then the result is marked green, otherwise it is marked red.

*Table 6.1*

| Mean Errors of the introduced metric 'ξ' per Hand for letter "kappa" | H1 | H2 | H3 | H4 |
|---|---|---|---|---|
| H1 | 1.9148 | 5.6079 | 3.4517 | 2.9744 |
| H2 | 5.9205 | 2.3377 | 4.6410 | 5.1033 |
| H3 | 3.5240 | 5.8539 | 2.8620 | 4.5486 |
| H4 | 3.1280 | 6.0830 | 4.5647 | 2.2223 |

| Mean Errors per Hand for letter "kappa" | H1 | | H2 | | H3 | | H4 | |
|---|---|---|---|---|---|---|---|---|
| | IDSC | + [45] | IDSC | + [45] | IDSC | + [45] | IDSC | + [45] |
| H1 | 52.2349 | 0.3929 | 72.7811 | 0.0449 | 66.5480 | 0.1436 | 60.3982 | 0.3808 |
| H2 | 72.8552 | 0.0109 | 47.3047 | 0.4683 | 61.4369 | 0.0568 | 74.7350 | 0.0096 |
| H3 | 66.6032 | 0.0314 | 61.3711 | 0.0455 | 52.3854 | 0.3068 | 66.2838 | 0.0286 |
| H4 | 64.2518 | 0.1983 | 74.7301 | 0.0206 | 66.2994 | 0.0535 | 48.6997 | 0.2357 |

*Table 6.2*

| Mean Errors of the introduced metric 'ξ' per Hand for letter "omega" | H1 | H2 | H3 | H4 |
|---|---|---|---|---|
| H1 | 3.8628 | 4.4319 | 4.8136 | 4.3468 |
| H2 | 5.5984 | 3.7810 | 4.1471 | 4.2074 |
| H3 | 6.9502 | 4.7370 | 2.8092 | 4.4478 |

| | | | | | | | | |
|---|---|---|---|---|---|---|---|---|
| H4 | | | 5.2737 | 4.8164 | 4.6113 | 4.2004 | | |

| Mean Errors per Hand for letter "omega" | H1 IDSC + [45] | | H2 IDSC + [45] | | H3 IDSC + [45] | | H4 IDSC + [45] | |
|---|---|---|---|---|---|---|---|---|
| H1 | 52.1593 | 0.3100 | 66.5975 | 0.0753 | 62.9115 | 0.1198 | 60.8586 | 0.1793 |
| H2 | 66.7478 | 0.0680 | 42.1727 | 0.3174 | 47.2358 | 0.2722 | 50.8315 | 0.2073 |
| H3 | 63.0522 | 0.1470 | 47.2105 | 0.3530 | 48.2460 | 0.3138 | 51.7896 | 0.2609 |
| H4 | 60.9506 | 0.1543 | 50.7885 | 0.2066 | 51.7506 | 0.1968 | 54.3374 | 0.1439 |

*Table 6.3*

| Mean Errors of the introduced metric 'ξ' per Hand for letter "epsilon" | H1 | H2 | H3 | H4 |
|---|---|---|---|---|
| H1 | 3.2294 | 4.7051 | 4.4754 | 4.7027 |
| H2 | | 4.6281 | 2.6699 | 3.5581 | 4.2574 |
| H3 | | 4.6692 | 3.4991 | 3.3061 | 4.0801 |
| H4 | | 4.5449 | 3.8123 | 4.1012 | 3.8047 |

| Mean Errors per Hand for letter "omega" | H1 IDSC + [45] | | H2 IDSC + [45] | | H3 IDSC + [45] | | H4 IDSC + [45] | |
|---|---|---|---|---|---|---|---|---|
| H1 | 52.6345 | 0.2962 | 70.0837 | 0.2081 | 62.5674 | 0.2225 | 64.9792 | 0.2447 |
| H2 | 70.1526 | 0.0986 | 47.9469 | 0.2104 | 57.5149 | 0.1526 | 62.2947 | 0.1148 |
| H3 | 62.6346 | 0.1287 | 57.4132 | 0.1852 | 46.9835 | 0.2067 | 59.3796 | 0.1491 |
| H4 | 66.2870 | 0.1762 | 62.2511 | 0.1555 | 59.2648 | 0.1690 | 50.8068 | 0.2238 |

**Tables 6.1-6.3**. Mean similarity/dissimilarity measures for all pairs of hands for the dataset of byzantine codices as obtained by the method introduced here and by two other methods. Verification of the ground truth classification is marked with green, while its violation is marked with red

The results for the data set of inscriptions and for letters "Alpha", "Sigma", "Omega", indicatively selected, are presented in tables 7.1-7.3 column-wise (i.e. similarity measurements of a hand with all others are presented as column vectors). This time, due to the larger number of the classifying hands and the large variability in alphabet symbols' shape the performance of the measures is marked by a 5-levels color bar which distinguishes the cases of ordering the correct group first, second, third, fourth and below fourth in the similarity ranking.

*Table 7.1*

Mean Errors 'ξ' per hand for "Alpha"

| | H1 | H2 | H3 | H4 | H5 | H6 | H7 | H8 | H9 | H10 |
|---|---|---|---|---|---|---|---|---|---|---|
| H1 | 7,1027 | 10,9131 | 9,8769 | 9,3292 | 8,2183 | 8,5123 | 8,2537 | 8,2948 | 10,8673 | 8,7028 |
| H2 | 12,4378 | 10,6681 | 13,4025 | 14,6639 | 14,2683 | 14,4788 | 11,5272 | 13,6154 | 16,2742 | 12,5559 |
| H3 | 10,5907 | 12,2772 | 7,9431 | 14,4333 | 12,9113 | 12,2790 | 9,9305 | 12,7618 | 13,9081 | 10,9243 |
| H4 | 9,2133 | 12,6222 | 13,3642 | 8,1717 | 8,1317 | 8,8181 | 9,7704 | 8,3030 | 10,4024 | 9,4267 |
| H5 | 8,2812 | 12,4517 | 12,1044 | 7,9122 | 7,2305 | 7,4876 | 8,6597 | 8,0458 | 10,8479 | 8,9845 |
| H6 | 8,4136 | 12,4225 | 11,1540 | 8,9123 | 7,5496 | 7,2472 | 8,2964 | 7,8141 | 11,3575 | 8,5109 |
| H7 | 8,4408 | 10,7950 | 9,0071 | 10,2573 | 9,0443 | 8,3328 | 7,4857 | 8,5211 | 10,6883 | 7,8137 |
| H8 | 8,7248 | 12,5633 | 12,3519 | 8,7731 | 8,4674 | 8,1779 | 9,0622 | 7,6927 | 13,7932 | 8,9721 |
| H9 | 11,7342 | 15,4650 | 14,0375 | 11,3666 | 11,8906 | 12,1459 | 11,0006 | 12,1072 | 10,5868 | 11,5090 |
| H10 | 9,0575 | 11,4288 | 10,6555 | 10,1398 | 9,5070 | 9,0405 | 8,1127 | 8,9630 | 11,1915 | 8,4630 |

Mean Errors IDSC, IDSC + [45] per hand for "Alpha"

| | H1 | | H2 | | H3 | | H4 | | H5 | |
|---|---|---|---|---|---|---|---|---|---|---|
| | IDSC | IDSC + [45] | IDSC | IDSC + [45] | IDSC | IDSC + [45] | IDSC | IDSC + [45] | IDSC | IDSC + [45] |
| H1 | 65,1990 | 0,2127 | 68,0053 | 0,2121 | 67,4789 | 0,2116 | 67,6560 | 0,2124 | 69,7811 | 0,2117 |
| H2 | 68,0396 | 0,1551 | 68,2306 | 0,1563 | 68,7371 | 0,1546 | 69,2564 | 0,1553 | 69,9397 | 0,1554 |
| H3 | 67,5283 | 0,2262 | 68,7455 | 0,2265 | 66,1170 | 0,2285 | 69,0948 | 0,2261 | 69,1876 | 0,2268 |
| H4 | 67,6822 | 0,1461 | 69,2453 | 0,1458 | 69,0568 | 0,1448 | 65,5193 | 0,1501 | 69,1766 | 0,1470 |
| H5 | 69,8114 | 0,1385 | 69,9384 | 0,1392 | 69,1688 | 0,1387 | 69,1979 | 0,1400 | 69,5641 | 0,1402 |
| H6 | 65,8774 | 0,1493 | 67,4082 | 0,1493 | 67,4146 | 0,1481 | 67,4434 | 0,1494 | 68,7610 | 0,1490 |
| H7 | 63,2111 | 0,2566 | 65,1032 | 0,2567 | 64,3608 | 0,2556 | 66,1165 | 0,2555 | 67,1467 | 0,2555 |
| H8 | 65,7094 | 0,1981 | 68,5155 | 0,1969 | 66,8299 | 0,1969 | 66,0095 | 0,1990 | 68,0029 | 0,1982 |
| H9 | 67,9782 | 0,2381 | 69,4086 | 0,2378 | 67,6000 | 0,2384 | 68,2748 | 0,2397 | 69,3665 | 0,2390 |
| H10 | 67,7672 | 0,2082 | 69,0622 | 0,2079 | 66,9224 | 0,2092 | 67,2095 | 0,2098 | 68,2418 | 0,2099 |

| | H6 | | H7 | | H8 | | H9 | | H10 | |
|---|---|---|---|---|---|---|---|---|---|---|
| | IDSC | IDSC + [45] | IDSC | IDSC + [45] | IDSC | IDSC + [45] | IDSC | IDSC + [45] | IDSC | IDSC + [45] |
| H1 | 65,8430 | 0,2120 | 63,1194 | 0,2110 | 65,6462 | 0,2117 | 67,9178 | 0,2123 | 67,6913 | 0,2113 |
| H2 | 67,4197 | 0,1551 | 65,0614 | 0,1533 | 68,4918 | 0,1534 | 69,3899 | 0,1550 | 69,0278 | 0,1534 |
| H3 | 67,4364 | 0,2258 | 64,3187 | 0,2251 | 66,8169 | 0,2254 | 67,5883 | 0,2271 | 66,8881 | 0,2268 |
| H4 | 67,4435 | 0,1456 | 66,0565 | 0,1420 | 65,9735 | 0,1460 | 68,2415 | 0,1472 | 67,1708 | 0,1457 |
| H5 | 68,7668 | 0,1386 | 67,0888 | 0,1364 | 67,7924 | 0,1382 | 69,3605 | 0,1395 | 68,2000 | 0,1389 |
| H6 | 65,2176 | 0,1495 | 62,2876 | 0,1481 | 65,1970 | 0,1490 | 67,8643 | 0,1489 | 67,0466 | 0,1477 |
| H7 | 62,3495 | 0,2574 | 57,6093 | 0,2605 | 62,4619 | 0,2563 | 65,9878 | 0,2551 | 64,6630 | 0,2547 |
| H8 | 65,2359 | 0,1981 | 62,4499 | 0,1964 | 61,3928 | 0,2035 | 65,9085 | 0,1994 | 64,2226 | 0,2002 |
| H9 | 67,8911 | 0,2369 | 65,9679 | 0,2327 | 66,1098 | 0,2381 | 66,0302 | 0,2569 | 66,5594 | 0,2386 |
| H10 | 67,0919 | 0,2078 | 64,6623 | 0,2059 | 64,2291 | 0,2107 | 66,5726 | 0,2104 | 64,4343 | 0,2136 |

Table 7.2
Mean Errors 'ξ' per hand for "Sigma"

| | H1 | H2 | H3 | H4 | H5 | H6 | H7 | H8 | H9 | H10 |
|---|---|---|---|---|---|---|---|---|---|---|
| H1 | 9,8088 | 13,8989 | 14,4567 | 15,0411 | 10,9800 | 11,3058 | 12,0799 | 13,1379 | 14,0800 | 10,5800 |
| H2 | 15,4189 | 11,7497 | 15,6708 | 14,8183 | 12,7033 | 15,0071 | 15,6650 | 14,6000 | 15,0971 | 13,9900 |
| H3 | 13,6628 | 13,5821 | 10,4973 | 13,7014 | 10,9125 | 11,7610 | 11,4886 | 10,7751 | 15,0129 | 12,2083 |
| H4 | 16,4456 | 14,7661 | 15,3622 | 11,9717 | 14,1883 | 16,5183 | 17,9544 | 14,4673 | 16,2900 | 15,0017 |
| H5 | 10,4473 | 11,6362 | 11,3300 | 12,6767 | 7,5940 | 9,7640 | 10,1520 | 10,4538 | 12,2143 | 9,8050 |
| H6 | 11,1423 | 13,4117 | 12,2183 | 15,1775 | 10,1435 | 10,2838 | 11,2239 | 13,2256 | 9,9280 | |
| H7 | 11,2533 | 13,3147 | 11,4022 | 15,2489 | 10,0372 | 10,2240 | 8,8945 | 10,8181 | 14,6183 | 9,9055 |
| H8 | 12,9408 | 12,7989 | 11,3244 | 13,4687 | 10,1628 | 11,4368 | 11,5736 | 10,3921 | 15,0138 | 12,0187 |
| H9 | 15,5233 | 15,0550 | 17,1158 | 16,4108 | 13,7585 | 15,0068 | 16,7058 | 16,3083 | 13,2400 | 13,3046 |
| H10 | 11,0975 | 13,3723 | 13,6050 | 15,2617 | 10,5433 | 10,6301 | 11,3300 | 13,1420 | 13,2630 | 9,8850 |

Mean Errors IDSC, IDSC + [45] per hand for "Sigma"

|  | H1 | | H2 | | H3 | | H4 | | H5 | |
|---|---|---|---|---|---|---|---|---|---|---|
|  | IDSC | IDSC + [45] | IDSC | IDSC + [45] | IDSC | IDSC + [45] | IDSC | IDSC + [45] | IDSC | IDSC + [45] |
| H1 | 75,1420 | 0,2741 | 72,7333 | 0,2737 | 73,9842 | 0,2734 | 74,8324 | 0,2760 | 73,6282 | 0,2738 |
| H2 | 72,7896 | 0,2628 | 68,5434 | 0,2607 | 69,2464 | 0,2626 | 73,9850 | 0,2626 | 69,0937 | 0,2627 |
| H3 | 74,0729 | 0,2793 | 69,2881 | 0,2795 | 71,0096 | 0,2765 | 75,3139 | 0,2791 | 69,7045 | 0,2796 |
| H4 | 74,8751 | 0,2565 | 73,9643 | 0,2534 | 75,2598 | 0,2530 | 72,4873 | 0,2601 | 74,9926 | 0,2535 |
| H5 | 73,7126 | 0,2534 | 69,1251 | 0,2532 | 69,7067 | 0,2532 | 75,0553 | 0,2530 | 69,9234 | 0,2562 |
| H6 | 73,0640 | 0,2680 | 69,4951 | 0,2672 | 70,5228 | 0,2670 | 73,7783 | 0,2681 | 69,9478 | 0,2676 |
| H7 | 72,7602 | 0,2903 | 66,1528 | 0,2918 | 66,8953 | 0,2919 | 74,8635 | 0,2896 | 67,1949 | 0,2914 |
| H8 | 71,7967 | 0,2978 | 66,8361 | 0,2982 | 68,0908 | 0,2981 | 72,7593 | 0,2978 | 67,9110 | 0,2982 |
| H9 | 74,3643 | 0,2441 | 70,0719 | 0,2437 | 70,8655 | 0,2436 | 73,4565 | 0,2453 | 70,2797 | 0,2441 |
| H10 | 74,4329 | 0,2842 | 69,1975 | 0,2844 | 69,7885 | 0,2846 | 74,7538 | 0,2847 | 68,9242 | 0,2851 |

|  | H6 | | H7 | | H8 | | H9 | | H10 | |
|---|---|---|---|---|---|---|---|---|---|---|
|  | IDSC | IDSC + [45] | IDSC | IDSC + [45] | IDSC | IDSC + [45] | IDSC | IDSC + [45] | IDSC | IDSC + [45] |
| H1 | 73,0134 | 0,2743 | 72,6382 | 0,2721 | 71,7197 | 0,2722 | 74,2386 | 0,2735 | 74,3126 | 0,2726 |
| H2 | 69,5110 | 0,2627 | 66,0737 | 0,2623 | 66,8119 | 0,2610 | 70,0023 | 0,2617 | 69,1245 | 0,2613 |
| H3 | 70,5743 | 0,2794 | 66,8580 | 0,2794 | 68,1063 | 0,2780 | 70,8359 | 0,2786 | 69,7464 | 0,2786 |
| H4 | 73,7765 | 0,2545 | 74,7526 | 0,2508 | 72,6953 | 0,2515 | 73,3561 | 0,2548 | 74,6584 | 0,2526 |
| H5 | 69,9835 | 0,2537 | 67,1601 | 0,2525 | 67,9149 | 0,2513 | 70,2287 | 0,2525 | 68,8868 | 0,2525 |
| H6 | 70,2199 | 0,2646 | 68,3471 | 0,2661 | 67,9993 | 0,2659 | 71,2328 | 0,2665 | 70,7151 | 0,2660 |
| H7 | 68,4339 | 0,2911 | 62,7622 | 0,2904 | 64,2730 | 0,2908 | 68,4288 | 0,2906 | 66,7194 | 0,2909 |
| H8 | 68,0386 | 0,2984 | 64,2154 | 0,2984 | 65,9496 | 0,2862 | 68,0435 | 0,2978 | 67,1034 | 0,2984 |
| H9 | 71,3175 | 0,2439 | 68,4152 | 0,2433 | 68,0871 | 0,2427 | 69,0955 | 0,2472 | 68,2290 | 0,2454 |
| H10 | 70,7978 | 0,2844 | 66,7174 | 0,2845 | 67,1585 | 0,2843 | 68,2390 | 0,2864 | 71,2965 | 0,2833 |

*Table 7.3*
*Mean Errors 'ξ' per hand for "Omega"*

|  | H1 | H2 | H3 | H4 | H5 | H6 | H7 | H8 | H9 | H10 |
|---|---|---|---|---|---|---|---|---|---|---|
| H1 | 3,7356 | 2,8201 | 3,4948 | 3,8354 | 5,9286 | 3,2944 | 4,1954 | 4,7504 | 3,9902 | 5,8628 |
| H2 | 4,2477 | 3,4474 | 3,6300 | 3,7362 | 4,3440 | 4,4316 | 3,5510 | 4,7045 | 3,0884 | 2,9587 |
| H3 | 5,0335 | 4,4689 | 3,4618 | 5,7344 | 6,1094 | 4,4685 | 3,8770 | 5,0970 | 3,8692 | 4,6570 |
| H4 | 5,2091 | 4,2847 | 3,9638 | 3,4989 | 3,5696 | 4,4917 | 4,7444 | 5,5603 | 4,1589 | 4,2485 |
| H5 | 4,8575 | 3,5417 | 3,3643 | 4,9861 | 3,1352 | 4,1879 | 3,9170 | 5,3785 | 3,0268 | 3,3296 |
| H6 | 5,1424 | 3,8263 | 3,9140 | 2,3806 | 3,7340 | 3,2766 | 4,1513 | 6,5179 | 3,5295 | 3,6623 |
| H7 | 4,9469 | 4,5413 | 4,6387 | 5,8829 | 5,5512 | 4,4005 | 3,6881 | 4,6214 | 3,9039 | 4,0556 |
| H8 | 5,0187 | 3,8016 | 4,4393 | 5,5462 | 3,7150 | 4,2350 | 4,6645 | 4,2316 | 3,4330 | 3,8911 |
| H9 | 4,0291 | 3,5288 | 3,4898 | 4,4313 | 4,4368 | 3,5780 | 3,7276 | 4,5590 | 2,5442 | 3,5020 |
| H10 | 3,7640 | 3,5203 | 3,7276 | 4,3452 | 4,7712 | 3,3514 | 3,2565 | 4,9613 | 2,6877 | 2,9987 |

*Mean Errors IDSC, IDSC + [45] per hand for "Omega"*

|  | H1 | | H2 | | H3 | | H4 | | H5 | |
|---|---|---|---|---|---|---|---|---|---|---|
|  | IDSC | IDSC + [45] | IDSC | IDSC + [45] | IDSC | IDSC + [45] | IDSC | IDSC + [45] | IDSC | IDSC + [45] |
| H1 | 64,4990 | 0,2207 | 63,8879 | 0,2196 | 68,6004 | 0,2125 | 75,6279 | 0,2090 | 73,3622 | 0,2073 |
| H2 | 63,8783 | 0,2250 | 60,2315 | 0,2273 | 68,0090 | 0,2123 | 74,3685 | 0,2104 | 72,0416 | 0,2098 |
| H3 | 68,5455 | 0,2175 | 67,9507 | 0,2105 | 66,3984 | 0,2317 | 71,7289 | 0,2324 | 68,6947 | 0,2340 |
| H4 | 75,5744 | 0,1213 | 74,2954 | 0,1166 | 71,7108 | 0,1378 | 71,7151 | 0,1570 | 70,2697 | 0,1499 |
| H5 | 73,2631 | 0,1441 | 71,9419 | 0,1406 | 68,6429 | 0,1631 | 70,2451 | 0,1728 | 75,8324 | 0,1476 |
| H6 | 67,5635 | 0,1792 | 67,2985 | 0,1751 | 68,2621 | 0,1823 | 73,0921 | 0,1856 | 71,2135 | 0,1825 |
| H7 | 62,8808 | 0,2588 | 59,4877 | 0,2629 | 68,2059 | 0,2422 | 75,1343 | 0,2390 | 72,3318 | 0,2385 |
| H8 | 65,4842 | 0,2552 | 59,0112 | 0,2628 | 67,7274 | 0,2480 | 73,5837 | 0,2478 | 70,4800 | 0,2490 |
| H9 | 65,8453 | 0,1888 | 60,8384 | 0,1950 | 68,6165 | 0,1794 | 73,7629 | 0,1795 | 71,1663 | 0,1803 |
| H10 | 68,1212 | 0,1884 | 63,9649 | 0,1915 | 69,7494 | 0,1847 | 73,5633 | 0,1873 | 71,3819 | 0,1861 |

**H6**        **H7**        **H8**        **H9**        **H10**

| | IDSC | IDSC + [45] | IDSC | IDSC + [45] | IDSC | IDSC + [45] | IDSC | IDSC + [45] | IDSC | IDSC + [45] |
|---|---|---|---|---|---|---|---|---|---|---|
| H1 | 67,6243 | 0,2172 | 62,8614 | 0,2215 | 65,5267 | 0,2140 | 65,8999 | 0,2160 | 68,1852 | 0,2155 |
| H2 | 67,3890 | 0,2191 | 59,4232 | 0,2308 | 59,0166 | 0,2273 | 60,4176 | 0,2287 | 64,0046 | 0,2255 |
| H3 | 68,2732 | 0,2248 | 68,1248 | 0,2088 | 67,6998 | 0,2096 | 68,5964 | 0,2086 | 69,7440 | 0,2124 |
| H4 | 73,0883 | 0,1338 | 75,0736 | 0,1138 | 73,5827 | 0,1176 | 73,7518 | 0,1165 | 73,5343 | 0,1222 |
| H5 | 71,1789 | 0,1550 | 72,2097 | 0,1381 | 70,3969 | 0,1432 | 71,1132 | 0,1423 | 71,3020 | 0,1459 |
| H6 | 69,2865 | 0,1799 | 67,0871 | 0,1748 | 68,0896 | 0,1720 | 68,2792 | 0,1732 | 69,7604 | 0,1754 |
| H7 | 67,1906 | 0,2507 | 58,8740 | 0,2613 | 59,3108 | 0,2599 | 60,6120 | 0,2612 | 64,2271 | 0,2576 |
| H8 | 68,1546 | 0,2519 | 59,2422 | 0,2631 | 55,5589 | 0,2609 | 57,6284 | 0,2654 | 61,6674 | 0,2618 |
| H9 | 68,3075 | 0,1851 | 60,5101 | 0,1953 | 57,5933 | 0,1961 | 60,7128 | 0,1923 | 62,1166 | 0,1955 |
| H10 | 69,7901 | 0,1879 | 64,1407 | 0,1912 | 61,6440 | 0,1925 | 62,1318 | 0,1943 | 65,8101 | 0,1808 |

*Color bar for the ranking of the correct group* 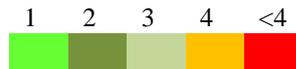

**Tables 6.1-6.3**. Mean similarity/dissimilarity measures for all pairs of hands for the dataset of inscriptions as obtained by the method introduced here and by two other methods. Ranking of the ground truth classification is marked according to the color bar.

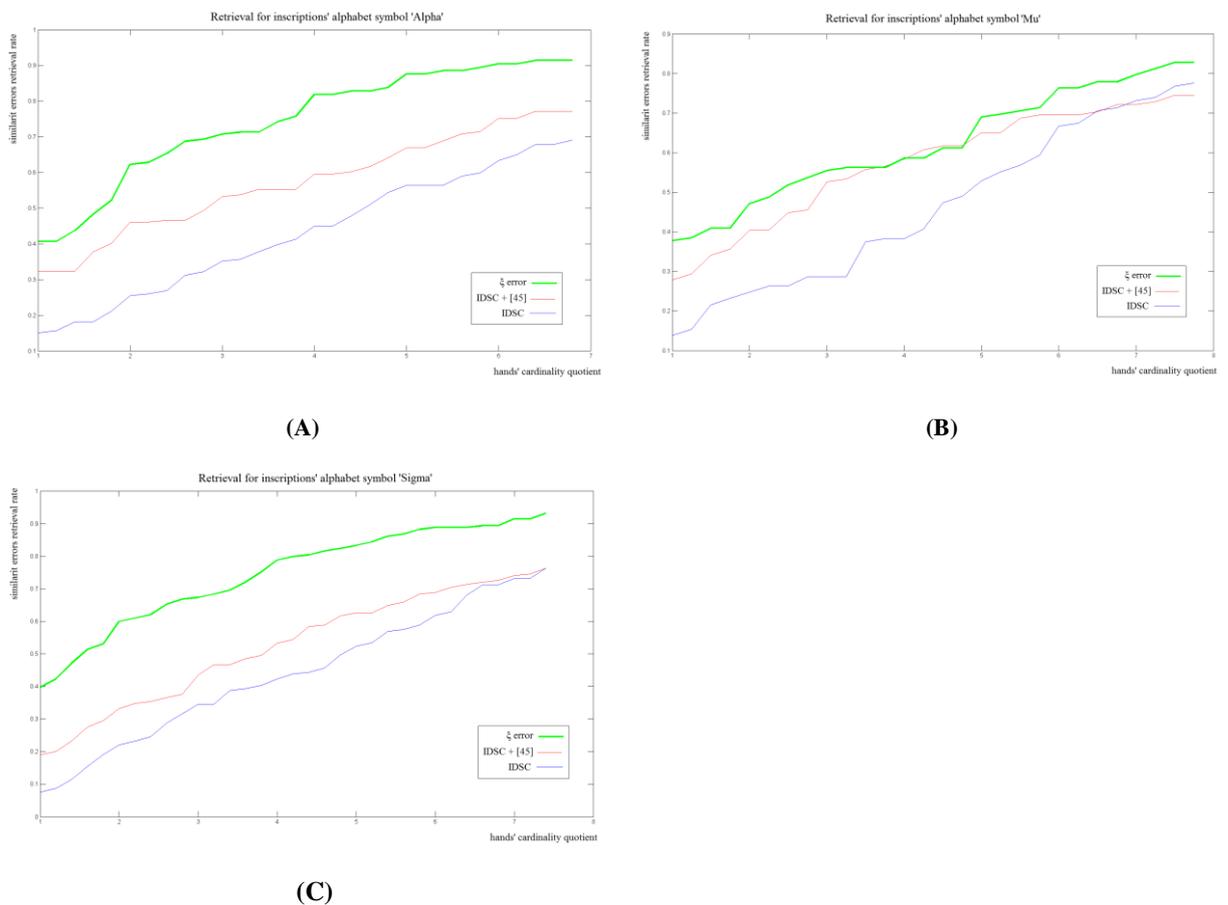

**(A)**

**(B)**

**(C)**

**Fig. 9** : The total retrieval results for letters 'Alpha', 'Sigma' and 'Mu' from the inscription's dataset.

Moreover, on the basis of the exhaustive pairwise similarity measurements, the retrieval rate achieved by the similarity metrics is computed in different quotients of the total population of alphabet symbols' shapes. Namely, for each letter realization separately we compute the ratio of the realizations that belong to the same hand with it in the best $k \cdot M$ similarity measurements, where $M$ is the hand's letters cardinality and $k$ is the retrieval's quotient. The total retrieval results for the introduced similarity error $\xi$ and the tested metrics of [44] and [45] are depicted in figures 9a-c.

## 9. Conclusion

In the present work, a novel methodology, which classifies curves into proper families according to their similarity, is introduced. It is assumed that the members of each such family remain in it, under any rotation, parallel translation and resize along axes $x$ and $y$ independently. This method has been applied to the problem of the automatic classification of ancient inscriptions and Byzantine codices according to their writer. Such a classification may become a really powerful, time-efficient and useful tool for accurate dating of ancient manuscripts. The introduced approach first extracts the contours of the alphabet symbols realizations appearing in each inscription or codex. Subsequently, a mathematical entity is introduced, which we call plane curvature and a number of fundamental propositions concerning it are stated and proved. In this way, it is demonstrated that, when a suitable, novel measure of plane curvature similarity is introduced, then there is a proper resize process that optimally minimizes the corresponding fitting error independently of rotation and parallel translation. Thus, a function minimization algorithm is applied, which offers optimal fit of any two curves, as far as plane curvature is concerned. Next, a Euclidean-type fitting error is employed, which takes rotation and parallel translation into consideration. Additionally, new similarity measures of two curves are defined, on the basis of the previous results. Statistical processing of the similarity measures values, by means of a new approach, offers both the number of distinct hands that have written these documents and the ensemble of inscriptions or codices belonging to each individual hand.

The introduced approach has been applied to 23 Byzantine codices and 46 ancient inscriptions of the Classical and the Hellenistic era. We would like to emphasize that the group of the authors consisting of Mathematics and Engineering specialists did not employ any reference manuscript, they had no idea about the number of distinct hands who had written the considered manuscripts and they did not have any related information whatsoever at their disposal. The developed system classified the 23 Byzantine codices in 4 different writers and the 46 inscriptions into 10 different hands. Prominent experts in Epigraphology, Archaeology and Classical Studies fully confirmed the classification offered by the system. We would like to emphasize that a scholar usually needs years of work to achieve such a classification.

The authors intend to expand the introduced methodology and to apply it to an even greater number of inscriptions and Byzantine codices, to tackle related open subjects and disputes and to extend the approach to other types of ancient documents.

## Appendix A – Calculation of formula (3.6)

Since $E$ is an isocontour of $F(x,y)$, then $\nabla F(x,y)$ is a vector normal to $E$ at $(x,y)$. Hence,

$$\vec{n} = \frac{\nabla F}{\|\nabla F\|} \Leftrightarrow n_x = \frac{\partial F/\partial x}{\|\nabla F\|}, \; n_y = \frac{\partial F/\partial y}{\|\nabla F\|}. \text{ Then, } C(x,y) = \nabla \cdot \vec{n} = \frac{\partial n_x}{\partial x} + \frac{\partial n_y}{\partial y} \Leftrightarrow$$

$$\Leftrightarrow C(x,y) = \frac{\partial}{\partial x}\left(\frac{\partial F/\partial x}{\|\nabla F\|}\right) + \frac{\partial}{\partial y}\left(\frac{\partial F/\partial y}{\|\nabla F\|}\right) = \frac{\dfrac{\partial^2 F}{\partial x^2}\|\nabla F\| - \dfrac{\partial F}{\partial x}\dfrac{\partial\|\nabla F\|}{\partial x}}{\|\nabla F\|^2} + \frac{\dfrac{\partial^2 F}{\partial y^2}\|\nabla F\| - \dfrac{\partial F}{\partial y}\dfrac{\partial\|\nabla F\|}{\partial y}}{\|\nabla F\|^2}$$

Expanding $\|\nabla F\| = \sqrt{\left(\dfrac{\partial F}{\partial x}\right)^2 + \left(\dfrac{\partial F}{\partial y}\right)^2}$ and using the matrix notation of Proposition 3 we have

$$\nabla\|\nabla F\| = \begin{bmatrix} \partial/\partial x \\ \partial/\partial y \end{bmatrix}\|\nabla F\| = \mathbf{H}(f)\frac{\nabla f}{\mu} \text{ and we can write the previous expansion of } C(x,y) \text{ in}$$

matrix notation as

$$C(x,y) = \frac{\nabla^T \nabla(f) - \dfrac{\nabla^T(f)}{\mu}H(f)\dfrac{\nabla(f)}{\mu}}{\mu} = \nabla^T(f)\frac{\left(I_2 \nabla^T \nabla - \nabla\nabla^T\right)(f)}{\mu^3}\nabla(f) =$$

$$= \nabla^T(f)\begin{bmatrix} 0 & -1 \\ 1 & 0 \end{bmatrix}\frac{\nabla\nabla^T(f)}{\mu^3}\begin{bmatrix} 0 & 1 \\ -1 & 0 \end{bmatrix}\nabla(f)$$

thus obtaining formula (3.6).

**Appendix B– Proof of Lemma1**

Total differential $dF$ is, by definition, written in the $(x, y)$ coordinates as

$$dF = \frac{\partial F}{\partial x} dx + \frac{\partial F}{\partial y} dy = \begin{bmatrix} dx & dy \end{bmatrix} \begin{bmatrix} \dfrac{\partial F}{\partial x} \\ \dfrac{\partial F}{\partial y} \end{bmatrix} \tag{4.6}$$

while, in $(\widetilde{x}, \widetilde{y})$, $dF$ is written as $dF = \dfrac{\partial F}{\partial \widetilde{x}} d\widetilde{x} + \dfrac{\partial F}{\partial \widetilde{y}} d\widetilde{y}$ . $\tag{4.7}$

But, after differentiating (4.2), we obtain $\begin{bmatrix} dx \\ dy \end{bmatrix} = \mathbf{A} \begin{bmatrix} d\widetilde{x} \\ d\widetilde{y} \end{bmatrix}$, $\tag{4.8}$

in which case, (4.6) becomes

$$dF = \begin{bmatrix} d\widetilde{x} & d\widetilde{y} \end{bmatrix} \mathbf{A}^T \begin{bmatrix} \dfrac{\partial F}{\partial x} \\ \dfrac{\partial F}{\partial y} \end{bmatrix} \tag{4.9}$$

Equating the right hand sides of (4.7) and (4.9) and taking into consideration the arbitrariness and independence of $(dx, dy)$ we obtain

$$\begin{bmatrix} \dfrac{\partial \widetilde{F}}{\partial \widetilde{x}} \\ \dfrac{\partial \widetilde{F}}{\partial \widetilde{y}} \end{bmatrix} = \mathbf{A}^T \begin{bmatrix} \dfrac{\partial F}{\partial x} \\ \dfrac{\partial F}{\partial y} \end{bmatrix} \tag{4.10}$$

since, (4.10) must hold for every differentiable $F$, (4.3) follows immediately.

Then, formula (4.4) for $\widetilde{\mu}$ follows immediately from the above formula (4.10) and its definition

$$\widetilde{\mu} = \sqrt{\begin{bmatrix} \dfrac{\partial \widetilde{F}}{\partial \widetilde{x}} & \dfrac{\partial \widetilde{F}}{\partial \widetilde{y}} \end{bmatrix} \begin{bmatrix} \dfrac{\partial \widetilde{F}}{\partial \widetilde{x}} \\ \dfrac{\partial \widetilde{F}}{\partial \widetilde{y}} \end{bmatrix}} .$$

Similarly, from the Hessian operator, we employ its definition $\tilde{\mathbf{H}} = \begin{bmatrix} \dfrac{\partial}{\partial \tilde{x}} \\ \dfrac{\partial}{\partial \tilde{y}} \end{bmatrix} \begin{bmatrix} \dfrac{\partial}{\partial \tilde{x}} & \dfrac{\partial}{\partial \tilde{y}} \end{bmatrix}$, together with (4.10) to obtain (4.5) immediately.

## Appendix C– Proof of Corollary 1

Relation (4.11) is written

$$\frac{dC}{C} = \frac{g_a da + g_b db}{\mu^2} \Leftrightarrow \frac{C + dC}{C} = \frac{\tilde{C}}{C} = e^{\frac{g_a da + g_b db}{\mu^2}} \qquad (C.1)$$

where the last relation comes from Taylor expansion of the exponential function keeping only the first order terms. If, now, we consider the new position as a starting one and we reapply the previous analysis, then we obtain $\dfrac{\tilde{C}_1}{\tilde{C}} = e^{\frac{g_a da^1 + g_b db^1}{\mu^2{}_1}}$. Evidently, $\dfrac{\tilde{C}_1}{\tilde{C}} = e^{\frac{g_a da + g_b db}{\mu^2} + \frac{g_a da^1 + g_b db^1}{\mu^2{}_1}}$. To

obtain the finite composite transformation, we integrate formula (C.1) to obtain:

$\ln\left( \dfrac{\tilde{C}(\tilde{x}, \tilde{y})}{C(x, y)} \right) = \displaystyle\int \frac{g_a da + g_b db}{\mu^2}$, from which (4.18) follows immediately.

## Appendix D– Proof of Proposition 5

In order to avoid unnecessary calculations of partial derivatives of the terms of (4.11), we will reevaluate the first order differential of the plane curvature function under the infinitesimal transformation (4.2) as $dC = d\left( \nabla \cdot \left( \dfrac{\nabla F}{\|\nabla F\|} \right) \right)$. So, action of (4.2) on $C(x, y)$ can be derived by the action of (4.2) on the gradient operator $\nabla$, which is given by (4.3). Thus, $d\nabla = d\mathbf{A}^T \nabla$, with $d\mathbf{A}$ given by (4.1) as $d\mathbf{A} = \begin{bmatrix} da & -dT \\ dT & db \end{bmatrix}$. Then, $dC = \nabla^T d\mathbf{A}\, \mathbf{n} + \nabla^T\left( \mathbf{l}\mathbf{l}^T \dfrac{d\nabla F}{\|\nabla F\|} \right) =$

$= \nabla^T d\mathbf{A}\, \mathbf{n} + \nabla^T\left( \mathbf{l}\mathbf{l}^T d\mathbf{A}^T\, \mathbf{n} \right)$, with $\mathbf{n} = \dfrac{\nabla F}{\|\nabla F\|}$ and $\mathbf{l} = \begin{bmatrix} 0 & -1 \\ 1 & 1 \end{bmatrix} \mathbf{n}$. Moreover, writing

$\nabla^T d\mathbf{A}\,\mathbf{n} = \nabla^T\left((\mathbf{l}\mathbf{l}^T + \mathbf{n}\mathbf{n}^T)d\mathbf{A}\,\mathbf{n}\right)$ and using the identities a) $\nabla^T \mathbf{n} = -\nabla^T \mathbf{l} = C(x,y)$, b)

$\mathbf{l}^T(\nabla \mathbf{l}^T) = \mathbf{l}^T(\nabla \mathbf{l}^T)\mathbf{n}\mathbf{n}^T = -C\mathbf{n}^T$, $\mathbf{n}^T(\nabla \mathbf{n}^T) = \mathbf{n}^T(\nabla \mathbf{n}^T)\mathbf{l}\mathbf{l}^T = -\mathbf{n}^T(\nabla \mathbf{l}^T)\mathbf{n}\mathbf{l}^T = C\mathbf{l}^T$ and c)

the fact that $d\mathbf{A} = \begin{bmatrix} da & 0 \\ 0 & db \end{bmatrix} + dT\begin{bmatrix} 0 & -1 \\ 1 & 0 \end{bmatrix}$ we have that

$$dC = \nabla^T(2\mathbf{l}\mathbf{l}^T d\mathbf{L}\,\mathbf{n} + \mathbf{n}\mathbf{n}^T d\mathbf{L}\,\mathbf{n}) = C(2\,\mathbf{l}^T d\mathbf{L}\,\mathbf{l} - \mathbf{n}^T d\mathbf{L}\,\mathbf{n}) \qquad (D.1)$$

where $d\mathbf{L} = \begin{bmatrix} da & 0 \\ 0 & db \end{bmatrix}$.

For the 2$^{nd}$ order differential and since the parameters $(a, b, T)$ are the independent variables of

the differentiation above, we have $d^2\mathbf{A} = 0$, thus obtaining

$$d^2\ln C = 2(2\,d\mathbf{l}^T d\mathbf{L}\,\mathbf{l} - d\mathbf{n}^T d\mathbf{L}\,\mathbf{n}) = -6d\mathbf{n}^T\mathbf{l}\,\mathbf{l}^T d\mathbf{L}\,\mathbf{n} = -6(\mathbf{l}^T d\mathbf{L}\,\mathbf{n})^2 \qquad (D.2)$$

Consequently, uniform scaling renders this 2$^{nd}$ order variation equal zero, thus implying that

$d^2\ln C$ only depends on the difference $da - db$, which is the differential eccentricity $a^2 - b^2$

at the origin ($a = 1, b = 1$). Actually, $d^2\ln C = -\dfrac{3}{2}\left(\mathbf{n}^T\begin{bmatrix} 0 & 1 \\ 1 & 0 \end{bmatrix}\mathbf{n}(da - db)\right)^2$. Since differential

$da + db$ of the scaling factor $ab$ at the origin is normal to the differential eccentricity and since

$da - db = \dfrac{\dfrac{da}{db} - 1}{\dfrac{da}{db} + 1}(da + db)$, $da - db$ and consequently $d^2\ln C$ only affect ratio $\dfrac{da}{db}$. This

fact makes 2$^{nd}$ order variations of $\ln C$ independent of the absolute size of the differential step

$(da, db)$; in fact these variations depend only on the direction of $(da, db)$.

## Appendix E– Proof of Proposition 1

In order to prove Proposition 1 we need to relate the form of $\zeta^C = \displaystyle\int_\Omega C(x,y)^2\,d\Omega$ with the form of $\varepsilon^C = \displaystyle\int_\Omega |C(x,y)|\,d\Omega$. Actually, exploiting the differential forms of $\varepsilon^C$ and $\zeta^C$,

$d\varepsilon^C = |C|\,d\Omega$ and $d\zeta^C = C^2 d\Omega = |C|\,d\varepsilon^C$ respectively, we have that the stationary

domains $\Omega$ of $\varepsilon^C$ and $\zeta^C$ coincide. Moreover, if, for any two domains $\Omega_1, \Omega_2$, we consider the path $Z : \Omega_1 \to \Omega_2$ and the supremum $C^Z$ of $| C(x, y) |$ on $Z$ we have that

$\zeta^C(\Omega_2) - \zeta^C(\Omega_1) \le C^Z \left( \varepsilon^C(\Omega_2) - \varepsilon^C(\Omega_1) \right)$. Thus, any collection of domains $\{\Omega_i\}$ ordered with respect to the corresponding values of $\varepsilon^C$ preserve this ordering when values of $\zeta^C$ are considered and vice versa. Namely, descending order of $\varepsilon^C$ values is retained by $\zeta^C$, while ascending order of $\zeta^C$ values is retained by $\varepsilon^C$.